\documentclass[twoside,11pt]{article}
\usepackage{jmlr2e-lyx}
\usepackage[ruled,vlined, linesnumbered,lined, boxed,commentsnumbered]{algorithm2e}
\usepackage{algorithm2e}
\usepackage{booktabs}
\usepackage{hyperref}
\usepackage{url}
\usepackage{amsmath}
\usepackage{amstext}
\usepackage{amssymb}
\usepackage{graphicx}
\usepackage[colorinlistoftodos]{todonotes}
\usepackage{latexsym}
\newtheorem{mydef}{Definition}
\newtheorem{Lemma}{Lemma}

\newtheorem{Theorem}{Theorem}

\title{Graph-Based Manifold Frequency Analysis for Denoising}

\author{\name Shay Deutsch \email shaydeut@usc.edu \\
       \addr Department of Computer Science\\
       University of Southern California\\
       Los Angeles, CA, USA
       \AND
       \name Antonio Ortega \email ortega@sipi.usc.edu  \\
       \addr Department of Electrical Engineering \\
       University of Southern California\\
       Los Angeles, CA, USA   
       \AND
       \name G\'{e}rard Medioni  \email medioni@usc.edu   \\
       \addr Department of Computer Science \\
       University of Southern California\\
       Los Angeles, CA, USA}

\begin{document}

\editor{}
\maketitle
\title{Graph-Based Manifold Frequency Analysis for Denoising}

\begin{abstract}We propose a new framework for manifold denoising based on processing in the graph Fourier frequency domain, derived from the spectral decomposition of the discrete graph Laplacian. Our approach uses the Spectral Graph Wavelet transform in order to perform non-iterative denoising directly in the graph frequency domain, an approach inspired by conventional wavelet-based signal denoising methods. 
We theoretically justify our approach, based on the fact that for smooth manifolds the coordinate information energy is localized in the low spectral graph wavelet sub-bands, while the noise affects all frequency bands in a similar way. 
Experimental results show that our proposed manifold frequency denoising (MFD) approach significantly outperforms the state of the art denoising methods, and is robust to a wide range of parameter selections, e.g., the choice of $k$ nearest neighbor connectivity of the graph. 
\end{abstract}

\begin{keywords}
Manifold Learning, Denoising, Graph Signal Processing, Spectral Graph Wavelets, Unsupervised Learning 
\end{keywords}

\section{Introduction}
Manifold learning has been proposed to extend linear approaches such as PCA in order to address the more general case where the data lies on a non-linear manifold. The existing manifold learning algorithms, e.g., \cite{Tenenbaum00}, \cite{LLE}, \cite{Belkin03}, \cite{HLLE} and \cite{Zhang05}, can provide effective tools to analyze high dimensional data with a complex structure when the data lies strictly on the manifold. However, in the presence of noise, i.e., when the observed data does not lie exactly on the manifold, the performance of these methods degrades significantly.  
Only a handful of methods have been suggested to handle noisy manifolds in a strictly unsupervised manner, e.g., \cite{MDhein}, \cite{Gongdenosie}, but  their main shortcoming is that they tend to over-penalize either the local or the global structure of the manifold.
In this paper, we address the manifold denoising problem using a graph-frequency framework called Manifold Frequency Denoising (MFD). 
Our approach is based on processing using Spectral Graph Wavelets (SGW), introduced by  \cite{Hammond}. Similar to time and frequency localization trade-offs provided by wavelets in regular signal domains, SGWs provide a trade-off between spectral and vertex domain localization. In the context of Machine Learning, this property allows us to overcome the limitations of existing manifold denoising methods, by providing a regularization framework in which the output denoised manifold is locally smooth without over-fitting or under-fitting the global manifold structure.    

\begin{figure}[t]
\vspace{.20in}
\includegraphics[trim=0cm 0cm 2cm 0cm, width=0.95\linewidth]{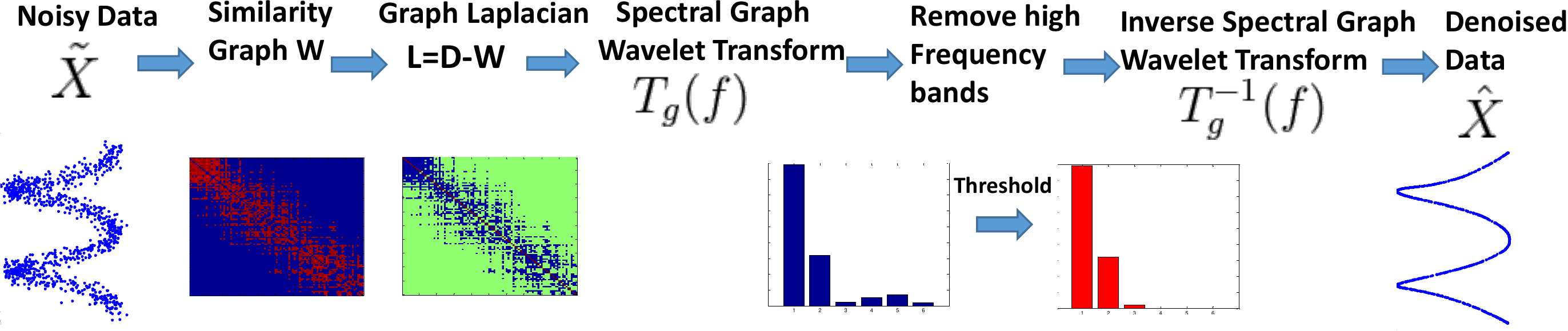}
\caption{Illustration of our method}
 \label{method_illustration}
\end{figure}

In our proposed framework (see Figure \ref{method_illustration} for an illustration) we build a graph where each vertex corresponds to one of the noisy observations, with edge weights between two vertices a function of the distance between the corresponding observations in the ambient space. Then, we apply the SGW to several graph-signals, where each graph-signal corresponds to one of the  dimensions, and assigns the scalar coordinate in that dimension to the corresponding  vertex. Thus, our graph has edge weights based on vector distances between observations, while denoising is applied to the observed coordinates in each dimension independently.         

In this paper, we theoretically justify our approach by showing that for smooth manifolds the coordinate signals also exhibit smoothness (i.e., the maximum variation across neighboring nodes is bounded). This is first demonstrated in the case of noiseless observations, by proving that manifolds with smoother characteristics lead to energy more concentrated in the lower frequencies of the graph spectrum. Moreover, it is shown that higher frequency wavelet coefficients decay in a way that depends on smoothness properties of the manifold. 
We then show that the manifold smoothness properties induce a similar decay characteristic on the spectral wavelets transform of the noisy signal. The effect of noise can be bounded for noisy graphs in which a large fraction of the edges in the noiseless graph remain connected.

Our experimental study also demonstrates that graph signal processing methods are effective for processing smooth manifolds, since in a graph signal defined based on these manifolds, most of the energy is concentrated in the low frequencies, making it easier to separate noise from signal information. To the best of our knowledge, MFD is the first attempt to use graph signal processing tools (\cite{GSP_irregular}) for manifold denoising. It is different from previous work for image denoising based on graph signal processing (\cite{image_denoise_laplacian}) in that the input data points are unstructured and our smoothness prior explicitly assumes that the original data lies on a smooth manifold.
 
  Another crucial aspect in manifold denoising is the efficiency and robustness of the process. Most of the current manifold denoising algorithms consist of iterative, global or semi global operations, which may also be sensitive to the parameter selection. In contrast, our denoising approach provides a fast and non-iterative process, with low computational complexity that scales linearly in the number of points for sparse data. Our approach is robust for a large range of parameter selections, and in particular, selection of $k$, the number of nearest neighbors used to construct the graph. In addition, our method does not require knowledge of the intrinsic dimensionality of the manifold.

Experimental results on complex manifolds and real world data, which includes motion capture and face expression datasets, demonstrate that our framework significantly outperforms the state of the art, so that, after denoising, it is possible to use current manifold learning approaches even for challenging complex smooth manifolds. Quantitatively, denoising using MFD significantly outperforms the state of the art denoising methods for a wide range of $k$ nearest neighbors selections both on synthetic and real datasets. In addition, our approach is shown to degrade gracefully as noise levels increase, while still preserving both local and global manifold structure. This paper is a significant extension to our recently published work on the subject (\cite{MFD}). In addition to extensive experimental validation beyond the one presented in \cite{MFD}, we provide a detailed theoretical analysis that characterizes the behavior of spectral graph wavelets both in the case of noiseless manifold observations and for the more general case where both the observations and the graph are noisy.

The rest of the paper is organized as follows: in Section \ref{sec:Related Work} we summarize the related work. In Section \ref{sec:Preliminaries} we introduce the notation and provide an overview of spectral graph wavelets.  Section \ref{sec:maintheory} presents our main theoretical results and Section \ref{sec:proposed} describes our new approach for manifold denoising. The experimental results are provided in Section \ref{sec:results} and in Section \ref{sec:conclusions} we conclude our work and suggest future work. 

\section{Related Work}
\label{sec:Related Work}
Denoising is very important for practical manifold learning, as most of the manifold learning approaches, e.g., Isomap (\cite{Tenenbaum00}), LLE (\cite{LLE}), LE (\cite{Belkin03}), LTSA (\cite{Zhang05}), and HLLE (\cite{HLLE}) assume that the data lies strictly on the manifold, and are known to be very sensitive to noise.
Thus a number of methods have been proposed to handle noisy manifold data. The state of the art methods include manifold denoising (MD, \cite{MDhein}), and  locally linear denoising (LLD, \cite{Gongdenosie}). Also related are statistical modeling approaches for manifold learning such as  Probabilistic non-linear PCA with Gaussian process latent variable models  (GP-LVM, \cite{Lawrence_gaussianprocess}), and its variants for manifold denoising (\cite{GaoCY08}).
  Among the state of the art methods, the method most related to our work is MD, which applies a diffusion process on the graph Laplacian, using an iterative procedure that solves differential equations on the graph. It is shown that each iteration in the diffusion process in MD is equivalent to the solution of a Tikhonov regularization problem on the graph (\cite{MDhein}). The main limitation of MD is over-smoothing of the data and sensitivity to the choice of $k$ nearest neighbor construction graph (as mentioned in \cite{MDhein}) especially at high noise levels. 

Our work is inspired by a classical approach for wavelet-based denoising (\cite{Donoho94idealspatial}), and its many extensions to image denoising (\cite{buades2005review}). Some recent work, (\cite{image_denoise_laplacian}), has explored graph-based techniques for image denoising for structured domains. However, the more general case of irregular domains is much less understood, and to the best of our knowledge, this work is the first attempt to explore spectral wavelets for manifold denoising on unstructured data. Spectral Graph Wavelets (SGW, \cite{Hammond}) provide us with an efficient tool to select spectral- and vertex-domain localization and are key component of our method (see Section~\ref{sec:Preliminaries} for more details). 

\section{Preliminaries}
\label{sec:Preliminaries}

\subsection{Notation}
\label{sec:Notation}

Consider a set of points  $\mathbf{x}=\left \{ \mathbf{x}_{i} \right  \}, \, i=1,...N, \mathbf{x}_{i} \in \mathbb{R}^{D}$, which are sampled from an unknown manifold $M$.  An undirected,  weighted graph $G=(V,E)$, is constructed over $\mathbf{x}$, where $V$ corresponds to the nodes and $E$  to the set of edges on the graph. The adjacency matrix  $\mathbf{W}= (w_{ij})$ consists of the weights $w_ {i,j} $ between node  $i $  and node $j$. 
In this work, the weights are chosen using the Gaussian kernel function 
\begin{equation}
\label{gaussian_weights}
W_{ij}=\left\{\begin{matrix}
\exp  \left (  \frac{  -||\mathbf{x}_{i}-\mathbf{x}_{j}||_{2}^{2} } {2\sigma_{D}^{2}}    \right )   & \mbox{if}\:  \mathbf{x}_{j} \in\text{kNN}(\mathbf{x}_{i}) \\ 
0 &  \mbox{otherwise} \end{matrix}\right.
\end{equation}
where is $(||\, ||)$ denotes the $L2$ distance between the points  $\mathbf{x}_{i}, \mathbf{x}_{j}$, kNN$(\mathbf{x}_{i})$ denotes the  $k$ nearest neighbors of $\mathbf{x}_{i}$, and $2\sigma_{D}^{2}$ are parameters used to construct the graph. The degree $d(i)$ of vertex $i$ is defined as the sum of weights of edges that are connected to $i$. In order to characterize the global smoothness of a function $ \mathbf{f}  \in \mathbb{R}^{N}$,  we  define its graph Laplacian quadratic form with respect to the graph as:\\
\begin{equation}
||\bigtriangledown \mathbf{f}|| ^{2}= \sum_{{i\sim j}}{w_{ij} (f(i)-f(j)})^{2} =  \mathbf{f}^{T}\mathbf{L} \mathbf{f}, 
\end{equation}
where $i \sim j$,  if $i$ and $j$ are connected on the graph by an edge, and $\mathbf{L}$ denotes the combinatorial graph Laplacian, defined as $\mathbf{L}=\mathbf{D}-\mathbf{W}$, with  $\mathbf{D}$ the diagonal degree matrix with entries $d_{ii}=d(i)$. 
The eigenvalues and eigenvectors of $\mathbf{L}$ are  $\lambda_1,\ldots,\lambda_N $ and $\mathbf{\phi}_{1},\ldots,\phi_N$, respectively.
The graph Fourier transform (GFT)  $ \hat{ f}$ of the function $f$ (which is a function over the vertices of the graph $G$),  is defined  as the expansion of $f$ in terms of the eigenvectors $\phi$ of the graph Laplacian, so that for frequency $\lambda_{l}$ we have:
\begin{equation}
\hat{f}(\lambda_{l}) = \sum_{i}{f(i) \phi_{l}^{*}(i)}.
\end{equation}
\subsection{Model Assumptions and Previous Results In Manifold Learning} 

We now recall the definitions of (i) the condition number $1/\tau$, (\cite{Niyogi08}), which provides an efficient measure to capture  both the local and global geometric properties of a manifold, and (ii) the geodesic covering regularity on the manifold (\cite{randomprojections}). For each $\mathbf{x}_{i} \in M$, where $M$ is a manifold, let $ T_{   \mathbf{x}_{i} }M$ and $ T_{   \mathbf{x}_{i} }^{\perp}M$ denote the tangent space and normal space to $M$, respectively. $B_D(\mathbf{x}_{i},r)$ is an open ball in $\mathbb{R}^{D}$ centered at $\mathbf{x}_{i}$ with radius $r$. The fiber of size $r$ at $\mathbf{x}_{i}$ is defined as $L_{r}\mathbf{x}_{i} = T_{\mathbf{x}_{i} }^{\perp}M\cap B_D(  \mathbf{x}_{i}  ,r)$. Given $\rho>0$, if $\rho <  \tau$, where $1/\tau$ is the condition number defined below, then $M\oplus \rho  $ is a disjoint union of its fibers, defined as (\cite{Minimax_manifold}):
\begin{equation}
M\oplus \rho = \underset{ \mathbf{x}_{i} \in M}{\cup} T_{ \mathbf{x}_{i}  }^{\perp}M\cap B_D(  \mathbf{x}_{i},\rho)
\end{equation}
\begin{figure}[htb]
\begin{minipage}[b]{1\linewidth}
\centering
\centerline{\includegraphics[width=12cm]{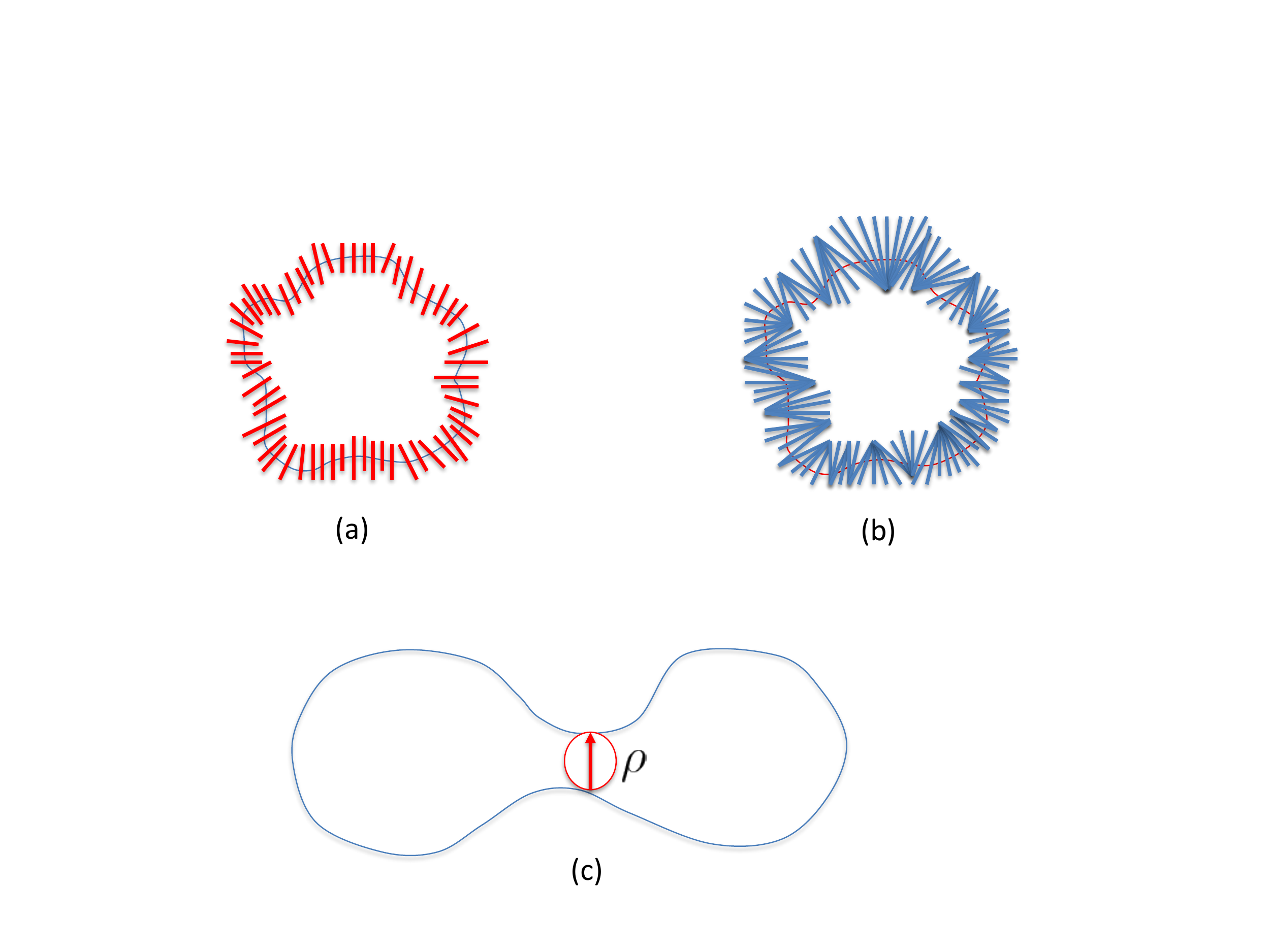}}
  \end{minipage}
\caption{Example of a smooth manifold. For the same manifold (a) shows the case  where the fibers do not extend beyond $\rho= \tau $, while (b) shows the case where the fibers intersect and extend beyond $\rho$. (c) is an example of a manifold with a condition number $1/\rho$.}
\label{condiiton_number_example}
\end{figure}

\begin{mydef} (Condition Number, \cite{Niyogi08})
The condition number of a manifold $M$ is the largest number $\rho$  such that each point in $M\oplus\rho$ has a unique projection onto $M$.
\label{condition_number}
\end{mydef}
Note that $\tau$ is small if $M$ is highly curved or close to self-intersections. Figure \ref{condiiton_number_example} provides an intuitive geometric illustration of the condition number of a manifold: when the fibers do not extend beyond $\rho= \tau $ (Figure \ref{condiiton_number_example}(a)), the projection from the fibers to the manifold is unique, which is not the case if the fibers extend beyond $\rho= \tau $ (Figure \ref{condiiton_number_example}(b)).
Given two points $\mathbf{x}_{i}, \mathbf{x}_{j}\in M$, let $d_M(\mathbf{x}_{i}, \mathbf{x}_{j})$ denote the geodesic distance between the points $\mathbf{x}_{i}, \mathbf{x}_{j}$. Also note that given a set of points $A$,  $|A|$ denotes the number of points in $A$. 
\begin{mydef} (Covering Number, (\cite{randomprojections}) Given $T>0$, the covering number $G(T)$ of a compact manifold $M$ is defined as the smallest number such that there exists a set $A$, with $G(T)= |A| $ , which  satisfies:
\begin{equation}
\underset{a\in A}{\min}\, d_{M}(\mathbf{x}_{i},a) \leq T
\end{equation}for all $\mathbf{x}_{i}\in M$.
\label{covering_number}
\end{mydef}

Note that given $T>0$, in order to achieve sufficiently dense covering of the manifold, the covering number $G(T)$ will depend on the volume and the intrinsic dimensionality of the manifold (\cite{randomprojections}). Also note that $G(T)$ provides the minimal number of points to cover the the manifold in resolution $T>0$.  

The following definition is often used to define localization properties in the graph domain:
\begin{mydef} (Shortest-path distance)
The shortest-path distance $d_{G}(m,n)$ between nodes $m$ and $n$ on the graph $G$ is the minimum number of edges for a path connecting $m$ and $n$, i.e.:
\begin{equation}
d_G(m,n) =  \underset{s}{\arg\min } \left \{ k_{1}, ...k_{s}  \right \}
\end{equation}
such that $k_{1}=m, k_{s}=n  $ and $ w_{ k_{r},  k_{r+1} } >0 $ for $1 \leq r <  s$.
\end{mydef}

\begin{mydef} 
Let $ \mathcal{N}(n, r)$ denote the set of vertex $n$'s neighbors in the graph which are $r$ hops away from $n$.
\end{mydef}

\subsection{Spectral Graph Wavelets} 
\label{sec:Spectral Graph Wavelets}

\begin{figure}[htb]
\begin{minipage}[b]{1\linewidth}
\centering
\centerline{\includegraphics[width=12cm]{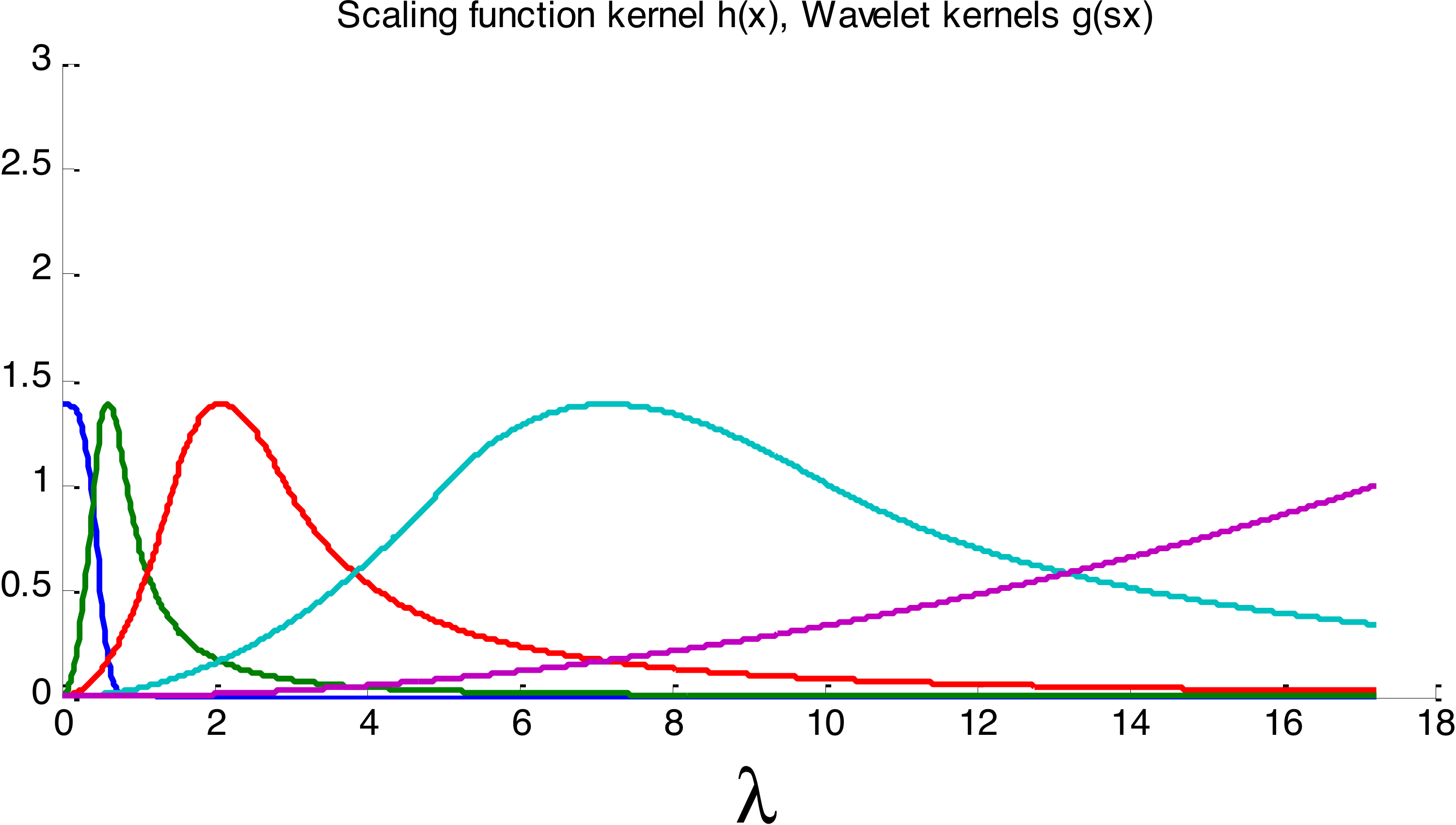}}
  \end{minipage}
\caption{Scaling function  (blue curve) and wavelet generating kernels g($\lambda$s) for different choices of scales s.}
\label{kerenel_wavelet_plot-crop.pdf}
\end{figure}

Spectral graph Wavelets (SGW)(\cite{Hammond}) define a scaling operator in the Graph Fourier domain introduced in Section 3.1 (see Figure \ref{kerenel_wavelet_plot-crop.pdf} for an illustration). SGWs are constructed using a kernel function operator  $T_g = g(\mathbf{L})$ which acts on a function $f$ by modulating each of its Fourier modes: 
\begin{equation}
\widehat{T_gf(\lambda_{l})}=g(\lambda_l)\widehat{f}(\lambda_{l}).
\end{equation} 
Given a function $f$, the wavelet coefficients take the form:
\begin{equation}
\Psi_{f}(s,n) =(T_{g}^{s}f(n))=\sum_{l=1}^{N} g(s\lambda_l)\widehat{f}(\lambda_{l})\phi_{l}(n).
\label{wavelet_coef_eqaulity}
\end{equation} 
The SGW can be computed with a fast algorithm based on approximating the scaled generating kernels by low order polynomials. 
The wavelet coefficients at each scale can then be computed as a polynomial of $\mathbf{L}$ applied to the input data.  When the graph is sparse, which is typically the case under the manifold learning model, the computational complexity scales linearly with the number of points, leading to a computational complexity of $O(N)$ (\cite{Hammond}) for an input signal $f \in  \mathbb{R}^{N}$. Including a scaling function corresponding to a low pass filter operation, SGWs map an input graph signal, a vector of dimension $N$, to $N(J+1)$ scaling and wavelet coefficients, which are computed efficiently using the Chebyshev polynomial approximation.  
Since $g$ is designed as a band pass filter in the spectral domain, graph wavelets will be localized in the frequency domain. The kernel function $g$ we use here is the same as the one defined in \cite{Hammond}. The kernel $g$ behaves as a band-pass filter satisfying the conditions $g(0) = 0 $  and $\lim_{x\rightarrow\infty  }g(x)=0$. A scaling function $h:  \mathbb{R}^{+}\rightarrow  \mathbb{R} $ acts as a lowpass filter which satisfies $h(0)>0$  and $h(x)\rightarrow 0$  when $x\rightarrow\infty$. Note that the scaling function helps ensure stable recovery of the original signal $f$ from the wavelet coefficients when the scale parameter $s$ is sampled at a discrete number of values $s_{j}$.

\section{Theoretical Results  }
\label{sec:maintheory}

\subsection{Overview}
\label{Overview}
We theoretically justify our framework, based on the following three main properties: 
\begin{description}
\item{i)} In manifolds that are smooth with respect to curvature and sampling rate, energy of a coordinate graph signal is concentrated in the low frequencies of the SGW. 
\item{ii)} The noise is flat in the SGW domain and the noise energy in each subband is proportional to the subband bandwidth. 
\item{iii)} Bounded noise perturbations in the data maintain the global structure of the graph associated to the true manifold, such that nodes that were connected in the noiseless graph are still connected in the noisy graph.  
\end{description}

 We assume that the noiseless points lie on a smooth or piecewise smooth manifold $M\in \mathbf{R}^{D}$. Let  $f_r()$ correspond to the values of all sampled points in dimension $r$, $r=1,...D $. In the noisy case, we assume that we are given a set of noisy points  $\tilde{f}_{r}(n)=f_{r}(n)+{\epsilon_{r}(n)}$, contaminated with Gaussian noise $\epsilon_{r}(n) \sim N(0,\sigma^{2})$  with zero mean and variance  $\sigma^{2}$. We assume the noise to be i.i.d. at each position and for each dimension $r$. In what follows, we describe the processing done for each of these signals and, unless required for clarity we drop the subscript $r$ and use  $f$ to denote the graph signal. We also assume that the graph signal $\mathbf{f}$ has non-zero mean $\bar{ \mathbf{f}}^2 \neq 0 $. 
 
We will characterize the behavior of the frequencies of the graph both in the noiseless and the noisy case. In the context of this paper, we call a 'noiseless graph', denoted by $G$, a graph that is constructed from a set of noiseless observations, hence the weights on the graph are noiseless. In contrast, we call a 'noisy graph', denoted by $\tilde{G}$, a graph constructed based on noisy observations, so graph weights and graph connectivity are subject to error due to noise. 
We will provide an analysis that will characterize the behavior of the graph Laplacian and the SGWs for both a noiseless graph $G$ and a noisy graph $\tilde{G}$. 

In  Section~\ref{subsec:smooth}  we set up the problem and consider the ideal scenario in which we are given a set of points that lie strictly on a smooth manifold.  We demonstrate that for the case of manifolds that are sufficiency smooth, as quantified by their condition number $1/\tau$ and geodesic covering number, the coordinates of each point on the graph change smoothly. This allows us to bound the variation of coordinate signals on  the graph as a function of the smoothness properties of the manifold in a way that shows that smoother manifolds will lead to coordinate signals with lower variation and thus lower graph frequencies. 

In Section \ref{subsec:noise-noiseless} we examine the case where the sampled points are noisy and the graph is noiseless. We demonstrate that the noise term affects all wavelet bands in a similar probabilistic way (similar distribution),  where the total energy of the noise in between adjacent scales differs by a logarithmic factor.

 In Section \ref{subsec:noisy-noisy} we address the most general case, where we have noisy observations as graph signals and a noisy graph $\tilde{G}$. Under a specific model of noise and sampling rate conditions (we assume that the noise is drawn uniformly on the normal to the manifold), we prove bounds on the decay of the energy of a noisy graph Laplacian and the noisy Spectral Graph Wavelets. Our results show that if the noise term is bounded by the smoothness properties  of the manifold (the condition number) and the sampling rate is sufficiently dense, then the energy of both the noisy graph Laplacian and the noisy spectral wavelets have decay properties  similar to those observed in the noiseless case.

\subsection{Noiseless Observations of Smooth Manifold }
 \label{subsec:smooth}

In the following lemma we will establish a connection between the smoothness of the manifold and the smoothness of the coordinate signal $f$. This lemma also motivates our choice of denoising each of the coordinate signals in the graph domain. By using a sufficiently high sampling rate, which depends on the smoothness properties of the manifold, we obtain that the points that are connected on the graph belong to the local neighborhood on the manifold, and that the corresponding coordinate signals vary smoothly.

\begin{Lemma} 
Consider a manifold $M$, with  a condition number $1/\tau$, which is sampled at a resolution of a geodesic covering number  $G(T)$, where also $\frac{T}{\tau}<1/4$. Let $\mathbf{f}$ be a graph signal which correspond to an arbitrary dimension $r$ of $M$. Then, for all $i, j\in G$  such that  $d_M( \mathbf{x}_{i}, \mathbf{x}_{j})<\delta $, where $ \delta = \min \left \{ \frac{\tau}{2}, 2CT  \right \} $ we have that: 
\begin{equation}
|f(i)-f(j) | \leq 4CT, 
\end{equation}
where $C$ is a constant, $ C \geq  1 $. 

\label{prop_manifold_graph_smooth}
 \end{Lemma}    

 {\textbf{Proof}} First note that for each $\mathbf{x}_{i}, \mathbf{x}_{j} \in M$ we have 
 \begin{equation}
|f(i)-f(j) | \leq  ||  \mathbf{x}_{i} - \mathbf{x}_{j}  ||  \leq d_M( \mathbf{x}_{i}, \mathbf{x}_{j})
\end{equation}
since the geodesic distance between two points on the manifold is always greater or equal to their euclidean distance. 
By Proposition 6.3  in \cite{Niyogi08}, we have that 
 \begin{equation}
d_{M}(\mathbf{x}_{i},\mathbf{x}_{j}) \leq  {\tau} -  {\tau}\sqrt{1- 2 ||\mathbf{x}_{i}-\mathbf{x}_{j} ||/ \tau} 
\end{equation}
 for  $\mathbf{x}_{i}, \mathbf{x}_{j}$ which obey $||\mathbf{x}_{i}- \mathbf{x}_{j}|| \leq \tau/2$. 
From this we can obtain:
 \begin{equation}
  {\tau} -  {\tau}\sqrt{1- 2 ||\mathbf{x}_{i}-\mathbf{x}_{j} ||/ \tau}  \leq   \frac{ 2||\mathbf{x}_{i}-\mathbf{x}_{j} || }{1 + \sqrt{1-  2||\mathbf{x}_{i}-\mathbf{x}_{j} ||/\tau }}\leq 2||\mathbf{x}_{i}- \mathbf{x}_{j}|| 
 \end{equation} 
We have that for all $ \mathbf{x}_{i}, \mathbf{x}_{j}$ for which  $d_M( \mathbf{x}_{i}, \mathbf{x}_{j})<\delta $ , where $ \delta = \min \left \{ \frac{\tau}{2}, 2CT  \right \} $,  $ C \geq  1$, since $||\mathbf{x}_{i}- \mathbf{x}_{j}|| \leq \delta$  and $T/\tau<1/4$, we have: 
 \begin{equation}
  ||\mathbf{x}_{i}- \mathbf{x}_{j}||  \leq d_{M}(\mathbf{x}_{i},\mathbf{x}_{j}) \leq  4CT
 \end{equation} 
and thus the inequality is obtained.                                                                               $\Box$         

Lemma \ref{prop_manifold_graph_smooth} shows that for a manifold which is sampled with a sufficient density, the manifold coordinate signal $\mathbf{f}$ changes smoothly, i.e., its local variation is bounded.  
Note that for a manifold with condition number $1/\tau $, the conditions  $T/\tau <1/4$, $||\mathbf{x}_{i}- \mathbf{x}_{j}|| \leq \tau/2 $ , limit the curvature and closeness to self-intersection. For these conditions, if  $\mathbf{x}_{i}, \mathbf{x}_{j} $ is such that $d_M(\mathbf{x}_{i}, \mathbf{x}_{j})<\delta $,  where then we obtain that the geodesic distance has the same order of the Euclidean distance, i.e., $d_M( \mathbf{x}_{i},\mathbf{x}_{j})\approx ||\mathbf{x}_{i}- \mathbf{x}_{j}||$. 
Defining 
\begin{equation}
\Delta(\frac{1}{\tau}, T) =4CT
\label{the_sampling_bound}
\end{equation} note that $\Delta(\frac{1}{\tau}, T)  $ decreases as $ T$ decreases. 

In the next lemma we bound the total variation of coordinate signals with 
respect to the graph as a function of the smoothness properties of the manifold. This lemma shows that smoother manifolds will lead to coordinate signals with lower variation and thus lower graph frequencies.  

\begin{Lemma}
Given a manifold $M$ with a condition number $ 1/\tau$, sampled at a resolution of a geodesic covering number $G(T)$, with the conditions of Lemma \ref{prop_manifold_graph_smooth} satisfied, and let $\mathbf{f}$ be a graph signal which correspond to an arbitrary dimension $r$ of $M$. Then the following inequality holds: 
\begin{equation}
||\bigtriangledown \mathbf{f}|| ^{2} \leq  \Delta(\frac{1}{\tau}, T)   \frac{ \lambda_{N} }{ C_{f }} \end{equation} where $C_{f} =\bar{ \mathbf{f}}^2$,  $\bar{ \mathbf{f}}^2 \neq 0 $ is the square of the mean of the graph signal  ${f}$. 
\label{Lemma_laplacian_bound}
\end{Lemma}
\textbf{Proof} Using the definition of the graph Laplacian we have 
\begin{align*} 
||\bigtriangledown \mathbf{f}|| ^{2} =  \mathbf{f}^{T}\mathbf{L} \mathbf{f}  = \sum_{ {i\sim j} } {w_{ij} (f(i)-f(j)})^{2} ,
\end{align*} 
next using normalization and the Cauchy-Schwartz inequality we obtain
\begin{equation} 
 \frac{ \sum_{{i\sim j}}{w_{ij} (f(i)-f(j)})^{2}  }  { ||f||^{2} }  \leq   \frac{ \sum_{{i\sim j}} {w_{ij} } ( f(i)-f(j) )^{2} }  { N \bar{f}^2} 
\label{eq:bound-var}
\end{equation} 
By applying Lemma \ref{prop_manifold_graph_smooth} with $T, C$ that obey its conditions, we can bound the coordinate signal difference terms in
(\ref{eq:bound-var}) for all vertices that are 1-hop neighbors on the graph: 
\begin{align*}
\frac{ \sum_{{i\sim j}} {w_{ij} } (f(i)-f(j))^{2} }  { N \bar{f}^2} \leq \frac{\Delta(1/\tau, T) \sum_{{i \sim j}} {w_{ij} }  } {  N\bar{f}^2} 
\label{eq:bound-var}
\end{align*}
where we used (\ref{the_sampling_bound}). 
Next summing over all vertices we get $\sum d_i < \sum d_{\max}$,  where $d_{\max}$ is the maximum degree and since   $d_{\max} < \lambda_{N}$  \cite{Lower_Bound_degree}, the Lemma is obtained. $\Box$ 

Essentially, the lemma states that if two manifolds with different  $\Delta(1/\tau, T$)  have the same Laplacian the coordinate signals corresponding to the smooth manifolds would have less variation $||\bigtriangledown \mathbf{f}|| ^{2}$ and thus more energy concentrated in the lower frequencies. This will be reflected in the SGW domain as well. Note that due to this property and the localization of spectral wavelets, the corresponding wavelet transform will be smooth. Also note that the localization property of SGW is achieved by an approximation of a $K$ degree polynomial, which leads to a $K$-hop localized transform in the spectral  wavelet  domain. A $K$ hop local neighborhood is a set of vertices that are within $K$ hops from a given vertex. Using Lemma 1 and Lemma 2, we now develop results for  the localization of Spectral Graph wavelet  coefficients of an arbitrary band $s$. We assume that  the  corresponding kernel function $g(s\lambda_{l})$ obeys the properties of the design in \cite{Hammond}. Our main assumption  on the kernel function  $g(s\lambda_{l})$ is that it is continuous and has a zero of integer multiplicity at the origin, i.e., $g(0)=0$ and $g^{r'}(0)=0$ for some integer $r'>0$. Also note that we assume that the graph is constructed as in (\ref{gaussian_weights}) but these results can also be applied using other types of distances, such as local tangent space distance.

\begin{Theorem} Given wavelet coefficients that were calculated from a smooth manifold with a condition number $1/\tau$ and a geodesic covering number $G(T)$ with the conditions of Lemma 1 satisfied, using a kernel function $g(s\lambda)$ which is non-negative in $[0,  \lambda_{\max}]$. Let $\mathbf{f}$ be a graph signal that corresponds to an arbitrary dimension $r$ of $M$. Then, the wavelet coefficients in a band $s$ obey
\begin{equation}
\sum_{n} |{\Psi_{ f } (s,n)} |^{2 } \leq    
s^2  \Delta(\frac{1}{\tau}, T)   \frac{ \lambda_{N} }{ C_{f }} C_{s},
\end{equation}
where, denoting $g_{s}^{'}(\lambda_{l})$ the derivative of the kernel function $g_{s}(\lambda_{l}) =g(s\lambda_{l})$, we have that for each $l$ 
\begin{equation}
g_{s}(\lambda_{l})=s g_{s}^{'}(c_{l}^{*}) \lambda_{l}
\end{equation}
for  $ c^{*}_{l} $ such that   $ s\lambda_{0} < c^{*}_{l}  <s \lambda_{l}$, and $C_{s} = \sum_{l} g_{s}'^{2}(c_{l}^{*})  \lambda_{l}$.

\label{Theorem_low_frequency}
\end{Theorem}
\textbf{Proof}
Observe that the following equality holds for any band $s$: 
\begin{equation}
\begin{split}
\sum_{n} |{\Psi_ { f}{(s,n)} }|^{2 } =\sum_{n}\sum_{l} g{(s\lambda_l)}\hat{f}(\lambda_l)\phi_l(n)  \sum_{l'} \overline{g{(s\lambda_{l'})}\hat{f}(\lambda_{l'})\phi_{l'}(n)   }
=   \sum_{l} |g{(s\lambda_l)}|^{2}| { \hat{f}(\lambda_l) } |^2   
\label{tight_frame}
\end{split}
\end{equation}
where in the first equality we used (\ref{wavelet_coef_eqaulity}). By construction of the kernel $g$, we have that it is continuous in $[\lambda_{\min}, \lambda_{\max}] $, and therefore also continuous in each interval $[0,s \lambda_{l}]$. By the mean value theorem for each $l$ there exists $c_{l}^{*}$ such that 

\begin{equation}
g(s\lambda_{l}) - g(0)= g_{s}^{'}(c_{l}^{*}) (s\lambda_{l}-s\lambda_{0}),
\end{equation}
 where $s \lambda_{0} < c_{*}  < s\lambda_{l}$. By the properties of spectral graph wavelets we have that  $g(0)=0$, whereas the first eigenvalue of the combinatorial Laplacian is $\lambda_{0} =0 $, and thus:   
\begin{equation}
g(s\lambda_{l})=s g_{s}^{'}(c_{l}^{*}) \lambda_{l}.
\label{mean_value}
\end{equation}
Using (\ref{mean_value}) and the Cauchy–Schwartz inequality we have: 
\begin{equation}
\label{applying_mean_value}
 \sum_{l} |g{(s\lambda_l)}|^{2}| { \hat{f}(\lambda_l) } |^2   =
 \sum_{l} s^{2}g_{s}'^{2}(c_{l}^{*}) \lambda_{l}^{2} | { \hat{f}(\lambda_l) } |^{2}  \leq
    s^{2}\sum_{l}g_{s}'^{2}(c_{l}^{*})   \lambda_{l}  \sum_{l}   \lambda_{l}   | { \hat{f}(\lambda_l) } |^{2}.
\end{equation}Finally, directly applying Lemma \ref{Lemma_laplacian_bound} we have

\begin{equation}
\begin{split}
 s^{2}\sum_{l} g_{s}'^{2}(c_{l}^{*})  \lambda_{l}  \sum_{l}   \lambda_{l}   | { \hat{f}(\lambda_l) } |^{2}  =  s^{2} \sum_{l}    \lambda_{l}g_{s}'^{2}(c_{l}^{*})  ||  \bigtriangledown  \mathbf{f}   ||^{2}  \\ 
 \leq   
  \Delta(\frac{1}{\tau}, T)   \frac{ \lambda_{N} }{ C_{f }} s^{2} \sum_{l} g_{s}'^{2}(c_{l}^{*})  \lambda_{l} = s^2  \Delta(\frac{1}{\tau}, T)   \frac{ \lambda_{N} }{ C_{f }} C_{s}
 \label{CS_equation}
 \end{split}
\end{equation}
where   
\begin{equation}
C_{s} = \sum_{l} g_{s}'^{2}(c_{l}^{*})  \lambda_{l}
\label{Cs_constant}
\end{equation}
and therefore the inequality is obtained.                                                                                                              $\Box$ \\

Thus, Theorem \ref{Theorem_low_frequency} shows that the spectral graph wavelets decay as a function of the smoothness properties of the manifold $ \Delta(\frac{1}{\tau}, T)$ and the scale $s$. Specifically, given two graph signals with the same coordinate energy $||\mathbf{f}||^{2}$, the bound $s^2  \Delta(\frac{1}{\tau}, T)   \frac{ \lambda_{N} }{ C_{f }} C_{s}$ would be smaller when the graph is smoother. \\
                                                                                                                                                                                                                                                                                                                                                                                                                                                                          \textbf{Remark 1:} Recall that increasing values of $s$ correspond to the lower frequencies. The results of Theorem \ref{Theorem_low_frequency} shows that for the terms depending on s in the bound  $s^2  \Delta(\frac{1}{\tau}, T)   \frac{ \lambda_{N} }{ C_{f }} C_{s} $ decay as  a function of $s$, i.e.,  $s^{2} C_{s}  \simeq s$, since the term $sC_{s}$ is an approximation of the admissibility condition (\cite{Hammond}). \\
 \textbf{Remark 2:} It is important to note that this result is guaranteed under sampling rates where the $k$ nearest neighbors on the graph are within a given geodesic distance on the manifold. Experimentally, the SGW transform and the suggested denoising approach demonstrate similar behavior and performance in cases where the local neighborhood includes points that are far in terms geodesic distance, as long as most of the nearest points on the graph belong to the true local neighborhood on the manifold.   \\
\textbf{Remark 3:} The theoretical results  stated above establish a new and interesting explicit connection between the smoothness in the graph domain and the graph frequencies via the curvature properties of the manifold (quantified by the condition number of the manifold). This result implies that the graph Laplacian and SGW bands will have more energy in the low frequencies  for manifolds with slowly changing or constant curvature. \\

\subsection{Noisy graph signals $\tilde{\mathbf{f}}$ over ${G}$} 
\label{subsec:noise-noiseless}
In this section we consider the case where the function is noisy and the Laplacian is noiseless. 
The next lemma shows that the noise term in the spectral wavelet coefficient has expectancy zero, and a variance that depends on the energy of the kernel filter in the spectral graph domain of the corresponding band $s$. Thus for the SGW designed using the kernel function suggested in \cite{Hammond} the variance of the noise term at arbitrary adjacent scales differs by a logarithmic factor. Note that the results obtained in this section are provided in terms of expectation, while in all the other sections the analysis provided is deterministic.

\begin{Lemma}
Consider the spectral wavelet coefficient obtained from a noisy function and noiseless Laplacian. 
Then we have that the expectation of the noisy wavelet coefficient term in scale $s$ is $E [ \Psi_ { \epsilon}(s,n) ] = 0$,  and
\begin{equation}
E[\sum_{i} |{\Psi_ { \epsilon}{(s,i)} }|^{2 } ] \leq  s^{2} \sigma^{2} C_{s}
\end{equation} where $C_{s}$ is given in (\ref{Cs_constant}).
\label{noise_noise_free}
\end{Lemma}
\textbf{Proof} First note that from the linearity of the spectral graph wavelets and since the noise $\epsilon(i) \sim N(0,\sigma^{2})$  is  i.i.d,  we immediately obtain: 
\begin{equation}
E [\Psi_ { \epsilon}(s,n )]= 0
\end{equation}
Next observe that by using (\ref{tight_frame}) and applying (\ref{applying_mean_value}) to the noisy graph signal we have 
\begin{equation}
\sum_{n} |{\Psi_ { \epsilon}{(s,n)} }|^{2 } =   \sum_{l} |g{(s\lambda_l)}|^{2}| { \hat{\epsilon}(\lambda_l) } |^2  = s^{2}   \sum_{l}  
(g_{s}^{'}(c_{l}^{*}) \lambda_{l})^{2}  | { \hat{\epsilon}(\lambda_l) } |^{2}
\end{equation}
 using the Cauchy-Schwartz inequality and Parseval's relation $\sum_{l}  | { \hat{\epsilon}(\lambda_l) } |^{2} =   { \sum_{n}|\epsilon(n) } |^{2} $  we obtain: 
\begin{align}
s^{2}   \sum_{l}  
(g_{s}^{'}(c_{l}^{*}) \lambda_{l})^{2}  | { \hat{\epsilon}(\lambda_l) } |^{2} 
 \leq s^{2}   \sum_{l}  (g_{s}^{'}(c_{l}^{*})  \lambda_{l})^{2}   \sum_{l}  | { \hat{\epsilon}(\lambda_l) } |^{2}   \\
 =   s^{2}   \sum_{l} (g_{s}^{'}(c_{l}^{*})  \lambda_{l})^{2}     { \sum_{n}|\epsilon(n) } |^{2}
\end{align}
Next taking the expectation we have 
\begin{equation}
\begin{split}
E[\sum_{n} |{\Psi_ { \epsilon}{(s,n)} }|^{2 } ]  \leq E [ s^{2}  \sum_{l} 
(g_{s}^{'}(c_{l}^{*})  \lambda_{l})^{2}     { \sum_{n}|\epsilon(n) } |^{2} ]   \\
= s^{2}    \sum_{l}  (g_{s}^{'}(c_{l}^{*})  \lambda_{l})^{2}  E[ {  \sum_{n}|\epsilon(n) } |^{2} ]  
= s^{2} \mbox C_{s} \sigma^{2}
\end{split}
\end{equation} $\Box$

The results above show that in the case of noisy function and noiseless Laplacian, the noise affects all bands in a way that is proportional to the bandwidth size, which, by construction, differs by a logarithmic factor between adjacent scales.

\subsection{Noisy graph signals $\tilde{\mathbf{f}}$ over $\tilde{G}$}
\label{subsec:noisy-noisy}

In this section, we provide an analysis for the case where both the function and the graph are noisy, under the assumption that the noisy points $ \left \{ \mathbf{\tilde{x}}_{i}  \right \}_{i=1}^{N},  \mathbf{\tilde{x}}_{i} = \mathbf{x}_{i}  + \xi_{i}$ are distributed uniformity on the normal to $M$, such that $ \xi_{i}$ is supported on $M\oplus \rho $  for all $\rho < \tau$. 
Note also that, in our experiments the distribution used was even more general (the noise was distributed in all directions and not restricted in all directions). It is also important to note that in practice the noiseless set of points $ \mathbf{x}_{i}$ is unknown, and the results obtained in this section for the noisy case assume that the manifolds are sampled sufficiently densely such that the conditions in Lemma  \ref{prop_manifold_graph_smooth}  are satisfied. We will denote $\mathbf{L}(G,W)$ and $\tilde {\mathbf{L}}(\tilde{G}, \tilde{W})$ the Laplacians constructed from the noiseless graph $G$ and the noisy graph $\tilde{G}$, respectively. $\tilde{d}(i)$ denotes the degree of the vertex $i$ in the noisy graph $\tilde{G}$. 
  The spectral wavelet coefficient which is constructed using a noisy function and a noisy Laplacian is denoted by $ \tilde{\Psi}_ {\tilde{f}}{(s,n)}  $. Let  $||\bigtriangledown \mathbf{\tilde{f}}||^2$ denote the graph Laplacian which is constructed from the noisy graph $\tilde{G}$, and a noisy graph signal $ \mathbf{\tilde{f}}$. 
In the following proofs we assume that the noiseless points were sampled under the conditions given in Section \ref{subsec:smooth}.
 
\begin{Lemma} 
Let $\tilde{X}= \left \{ \mathbf{\tilde{x}}_{i}  \right \}_{i=1}^{N}$ be a set noisy points, and let $ \mathbf{ \tilde{f} }$ be a noisy graph signal that corresponds to an arbitrary dimension $r$ of  $\mathbf{\tilde{x}}_{i} $. Assume that  the conditions of Lemma \ref{prop_manifold_graph_smooth} are satisfied for the points $ \left \{ \mathbf{x}_{i}  \right \}_{i=1}^{N} $, then we have that the following inequality holds:  
\begin{align}  
||\bigtriangledown \mathbf{\tilde{f}}|| ^{2}   \leq        \Delta(\frac{1}{\tau}, T + q(\xi)   )   \frac{ \lambda_{N} + d_{\max}^{(2)}  }{ C_{ \tilde{f} }}   
\end{align}where 
\begin{equation}
q(\xi)  =2 \,\min_{i \in \tilde{G}}  \max \left \{  \xi_{i},  \frac{4C^{2}T^{2}}{\tau }    \right \},
\end{equation}
\begin{equation}
d_{\max}^{(2)} = \max_{i \in {G}}   \left \{  | \mathcal{N}(i, 2) |   \right \}   , 
\label{k_G_constrution_parameter}
\end{equation}
and $C_{\tilde{f}} =\bar{\tilde{\mathbf{f}}}^2$ is the square of the mean of the noisy graph signal  ${\tilde{f}}$. 
\label{Noisy_laplacian_bound}
\end{Lemma}
\textbf{Proof}: Using the definitions of the graph Laplacian with the noisy set of points  $\tilde{X}$ we have 
\begin{align} 
||\bigtriangledown \mathbf{\tilde{f}}|| ^{2} =  \tilde{f}^{T}\mathbf{\tilde{L}} \tilde{f}  = \sum_{ {i\sim j} } {\tilde{w}_{ij} (\tilde{f}(i)-\tilde{f}(j)})^{2} 
\label{noisy_laplacian_eq}
\end{align}
Let $\mathbf{\tilde{x}}_{i} \in {\tilde{X}} $ and denote $\tilde{B} = B_{D}(\mathbf{\tilde{x}}_{i} ,\tilde{\delta})$ for some $\tilde{\delta} > 0 $. By Lemma  4.1 in \cite{Niyogi08} we have that if $ v \in B_{D}( \mathbf{{x}}_{i},\tau) \cap  T_{ \mathbf{{x}}_{i}}^{\perp}M \cap  B_{D}( \mathbf{{x}}_{j},\delta)$,  $ \delta \geq T$ and $ \mathbf{{x}}_{j} \notin  B_{D}( \mathbf{{x}}_{i},\delta) $  then $  || \mathbf{{x}}_{i} - v || <  \delta^{2}/\tau $. Now take $\tilde{\delta}= \delta +    \frac{4C^{2}T^{2}}{\tau }$. First observe that $ (B_{D}(\mathbf{x}_{i}, \delta) \cap \tilde{B})  \neq \emptyset  $, since $M$ is  sampled at a resolution of geodesic cover  $G(T)  $. Applying Lemma 4.1 from \cite{Niyogi08} in our case with $v = \mathbf{\tilde{x}}_{i} $, in the worst case scenario,  if  $ \mathbf{{x}}_{j} \notin  B_{D}( \mathbf{{x}}_{i},\delta) $, but $ \mathbf{{x}}_{j} \in  B_{D}( \mathbf{{x}}_{i}, \tilde{\delta}) $ , we obtain that $  || \mathbf{{x}}_{i} -  \mathbf{\tilde{x}}_{i} || \leq    \frac{4C^{2}T^{2}}{\tau }$.  
For all $i,j \in \tilde{G}$  such that $d_M( \mathbf{\tilde{x}}_{i}, \mathbf{\tilde{x}}_{j}) < \tilde{\delta}$, we have that: 

\begin{equation}
|\tilde{f}(i)-\tilde{f}(j) | \leq  ||  \mathbf{\tilde{x}}_{i} -  \mathbf{\tilde{x}}_{j}  ||  \leq   ||  \mathbf{{x}}_{i} +\xi_{i}-  \mathbf{{x}}_{j}  - \xi_{j} || \leq  \\
||  \mathbf{{x}}_{i} -  \mathbf{{x}}_{j}   || +  2 \, \max\left \{ \xi_{i},  \xi_{j}\right \} \label{ineq_noisy_function}
\end{equation}
where $\xi_{i}$ denotes the noise corresponding to $ \mathbf{{x}}_{i}$. \\
Denoting
\begin{equation}
q(\xi)  =2 \, \min_{i\in \tilde{G}}  \max \left \{  \xi_{i},     \frac{4C^{2}T^{2}}{\tau }   \right \}
\end{equation}
we have that $\forall i\in G$
\begin{equation}
||\xi_{i}|| \leq  ||  \mathbf{\tilde{x}}_{i} -  \mathbf{x}_{i}  ||   \leq q(\xi) /2. 
\end{equation}
By using normalization and the Cauchy-Schwartz inequality as in Lemma \ref{Lemma_laplacian_bound}, and then applying inequality (\ref{ineq_noisy_function}) in (\ref{noisy_laplacian_eq}) we obtain  
\begin{equation}
\frac{  \Delta(\frac{1}{\tau}, T + q(\xi)   ) \sum_{i \sim j} { \tilde{w}_{ij}}  }{  ||\tilde{f}||^{2} }   \leq  \frac{  \Delta(\frac{1}{\tau}, T + q(\xi)   ) \sum_{i \sim j} { \tilde{w}_{ij}}  }{  N \bar{\tilde{f}}^2 }  
\end{equation}
For the sampling conditions of Lemma 1, we obtain that $ \delta  \geq   \frac{4C^{2}T^{2}}{\tau }$ and thus every vertex in $\tilde{G}$ has at most $d_{\max} + d_{\max}^{(2)}  $ edges, where $d_{\max}^{(2)}$ is defined as: 
\begin{equation}
d_{\max}^{(2)} = \max_{i \in {G}}   \left \{  | \mathcal{N}(i, 2) |   \right \}  
\label{max_two_hopes}
\end{equation}
i.e., $d_{\max}^{(2)}$ is the maximum number of vertices on the graph $G$ that are two hops away from an arbitrary vertex. 
Summing over all vertices we get that $\sum \tilde{d}(i) < \sum (d_{\max} +d_{\max}^{(2)}) $ and since $d_{\max}<\lambda_{N}$, the lemma is obtained. $\Box$ 

Thus in the presence of noise the decay in the energy of the noisy Laplacian depends on an additional parameters the distribution of the vertex degrees in the graph $d_{\max}^{(2)} $ (that depends on the smoothness properties of the graph) and the bounded noise term $ q(\xi)$. Therefore bounded noise leads to bounded increase increase in variation. As will be shown in the next theorem, these properties will control the decay of SGWs frequency bands as well, modulated by power two of the scale $s$ and and the constant $\tilde{C}_{s}$ which is essentially an estimation of the energy of the kernel filter in a band of scale $s$.

\begin{Theorem}
Given a set of noisy points $\tilde{X}= \left \{ \mathbf{\tilde{x}}_{i}  \right \}_{i=1}^{N}$, and a noisy graph signal $ \mathbf{\tilde{f}}$ that corresponds to an arbitrary dimension $r$ of  $\mathbf{\tilde{x}}_{i}$. Assume that the conditions of Lemma \ref{prop_manifold_graph_smooth} are satisfied for the points $\left \{ \mathbf{x}_{i}  \right \}_{i=1}^{N}$. Then, the noisy spectral wavelets  in a band $s$, calculated using  the kernel function $g(s\tilde{\lambda})$ which is non-negative in $[0,  \tilde{\lambda}_{N}] $, obey: 

\begin{equation}
 \sum_{n} | \tilde{\Psi}_ {\tilde{f}}{(s,n)}|^{2}    \leq s^{2}  \Delta(\frac{1}{\tau}, T + q(\xi)   )   \frac{ \lambda_{N} + d_{\max}^{(2)}  }{ C_{ \tilde{f} }}  \tilde{C_{s}}
 \label{thorem1_inequality}
\end{equation}
where for each $l$ we have that 
\begin{equation}
g_{s}(\tilde{\lambda}_{l})=s g_{s}^{'}(\tilde{c}_{l}^{*}) \tilde{\lambda}_{l}
\end{equation}for  $ \tilde{c}^{*}_{l} $ such that   $ s\tilde{\lambda}_{0} < \tilde{c}^{*}_{l}  <s \tilde{\lambda}_{l}$,  $ \tilde{C_{s}} = \sum_{l} g_{s}'^{2} \tilde{\lambda}_{l} ( \tilde{c}_{l}^{*}) $, and $d_{\max}^{(2)} $ is the maximum degree of two hops on the graph G which is defined in (\ref{max_two_hopes}). 
  \label{Noisy_Theorem}
\end{Theorem}
\textbf{Proof}
We first observe that the following equation holds using the definitions for the noisy wavelet coefficients: 

\begin{equation}
\begin{split}
\sum_{n} | \tilde{\Psi}_ {\tilde{f}}{(s,n)}|^{2}  =   \sum_{l} |g{(s\tilde{\lambda_l})}|^{2}| { \hat{\tilde{f}}(\lambda_l) } |^2   
\end{split}
 \label{noise_noiseless_eq}
\end{equation}
where $g_s(\tilde{\lambda})$ denotes the square of the kernel function applied on the domain of the noisy eigenvalues $\tilde{\lambda}_{l}$. 
Using similar arguments as in Theorem 1 (by applying the mean value theorem to the function$g_{s}(\tilde{\lambda}_{l})$) and using the Cauchy-Schwartz inequality we obtain: 
\begin{equation}
\begin{split}
 \sum_{l} g_{s}{(\tilde{\lambda}_l) } | { \hat{\tilde{f}}(\lambda_l) } |^2 \,   = s^{2} \sum_{l} g_{s}'^{2} ( \tilde{c}_{l}^{*}) \tilde{\lambda}_{l}^{2} | { \hat{\tilde{f}}(\tilde{\lambda_l}) } |^{2}   \\
 \leq s^{2} \sum_{l} g_{s}'^{2}( \tilde{c}_{l}^{*})\tilde{\lambda}_{l} \sum_{l}   \tilde{\lambda}_{l} | { \hat{\tilde{f}}(\tilde{\lambda_l}) } |^{2} =  s^{2} \sum_{l}g_{s}'^{2}( \tilde{c}_{l}^{*})\tilde{\lambda}_{l} ||\bigtriangledown \mathbf{\tilde{f}}|| ^{2}
\end{split}
\end{equation}
finally using Lemma \ref{Noisy_laplacian_bound} we obtain 
\begin{equation}
\begin{split}
s^{2} \sum_{l} g_{s}'^{2}( \tilde{c}_{l}^{*}) ||\bigtriangledown \mathbf{\tilde{f}}|| ^{2} \leq  s^{2} \sum_{l} g_{s}'^{2} ( \tilde{c}_{l}^{*}) \tilde{\lambda}_{l}  \Delta(\frac{1}{\tau}, T + q(\xi)   )   \frac{ \lambda_{N} + d_{\max}^{(2)}   }{ C_{ \tilde{f} }}  \\
= s^{2}  \Delta(\frac{1}{\tau}, T + q(\xi)   )   \frac{ \lambda_{N} +  d_{\max}^{(2)}  }{ C_{ \tilde{f} }}   \tilde{C_{s}}
\end{split}
\end{equation}
where $ \tilde{C_{s}}  = \sum_{l} g_{s}'^{2} ( \tilde{c}_{l}^{*}) \tilde{\lambda}_{l}  $     $\Box$
\\\\
\textbf{Remark 1:} The result of Theorem \ref{Noisy_Theorem} implies that, for bounded noise (which depends on the condition number of the manifold),  a large fraction of the edges in the noiseless Laplacian  remain connected in the noisy Laplacian. Thus we obtain that the decay of the noisy spectral graph wavelets is similar to the decay of the noiseless spectral graph wavelets. Note that the number of new edges introduced in the noisy graph depends on an additional parameter, $d_{\max}^{(2)}$, which bounds the number of edges that can be added to an arbitrary vertex in the noisy graph. This result can also be understood by the way our graph is constructed. Bounded perturbations of the data maintain a graph topology that is similar to the unknown, noiseless graph. This property,  combined with the assignment of the manifold coordinates as graph signals, preserve a measure of locality, which is needed in learning the structure of manifolds. \\

\textbf{Remark 2:} Since $d_{\max}^{(2)} $ is a parameter that depends on the smoothness properties of the graph, then Lemma \ref{Noisy_laplacian_bound} and Theorem  \ref{Noisy_Theorem} can be interpreted as follows: if the manifold is sampled sufficiently densely (when $ \Delta(\frac{1}{\tau}, T) $ is sufficiently small), then the noise becomes less significant and the decay in the noisy case is similar to the noiseless one. \\
\textbf{Remark 3:} Note that  if the noise is zero, then we are reduced to the noise-free case, as one would have expected.\\
\textbf{Remark 4:}  In the case where $\mathbf{L}=\tilde{\mathbf{L}}$ (the Laplacian is noise-free and thus only the function is noisy) we obtain the following bound for the energy in the noisy spectral wavelets in a band $s$:  
\begin{align*}  
||\bigtriangledown \mathbf{\tilde{f}}|| ^{2}   \leq        \Delta(\frac{1}{\tau}, T + q(\xi)   )   \frac{ \lambda_{N} }{ C_{ \tilde{f} }}   
\end{align*}
while on the other hand we can observe that the noisy-noisy case adds the term $   \Delta(\frac{1}{\tau}, T + q(\xi)   )   \frac{  d_{\max}^{(2)}  }{ C_{ \tilde{f} }}$.

\section{Proposed Approach}
\label{sec:proposed}

We now describe our approach for manifold denoising. As before, $f_r()$ correspond to the values of all sampled points in dimension $r$. Denoising is performed independently for each $f_r()$. Following the noise model introduced in \ref{Overview}, the goal is to provide an estimate $\hat{f}_r(i)$ of the original coordinates $f_r(i)$ given $\tilde{f}_r(i)$ for each $r$ and for all $i$. The reconstructed manifold points are estimated by constructing ${\hat{\mathbf{x}}_i}$, which is based on $\hat{f}_{r}(i)$. 

\begin{figure}[htb]
\begin{minipage}[b]{1\linewidth}
  \centering
  \centerline{\includegraphics[width=14cm]{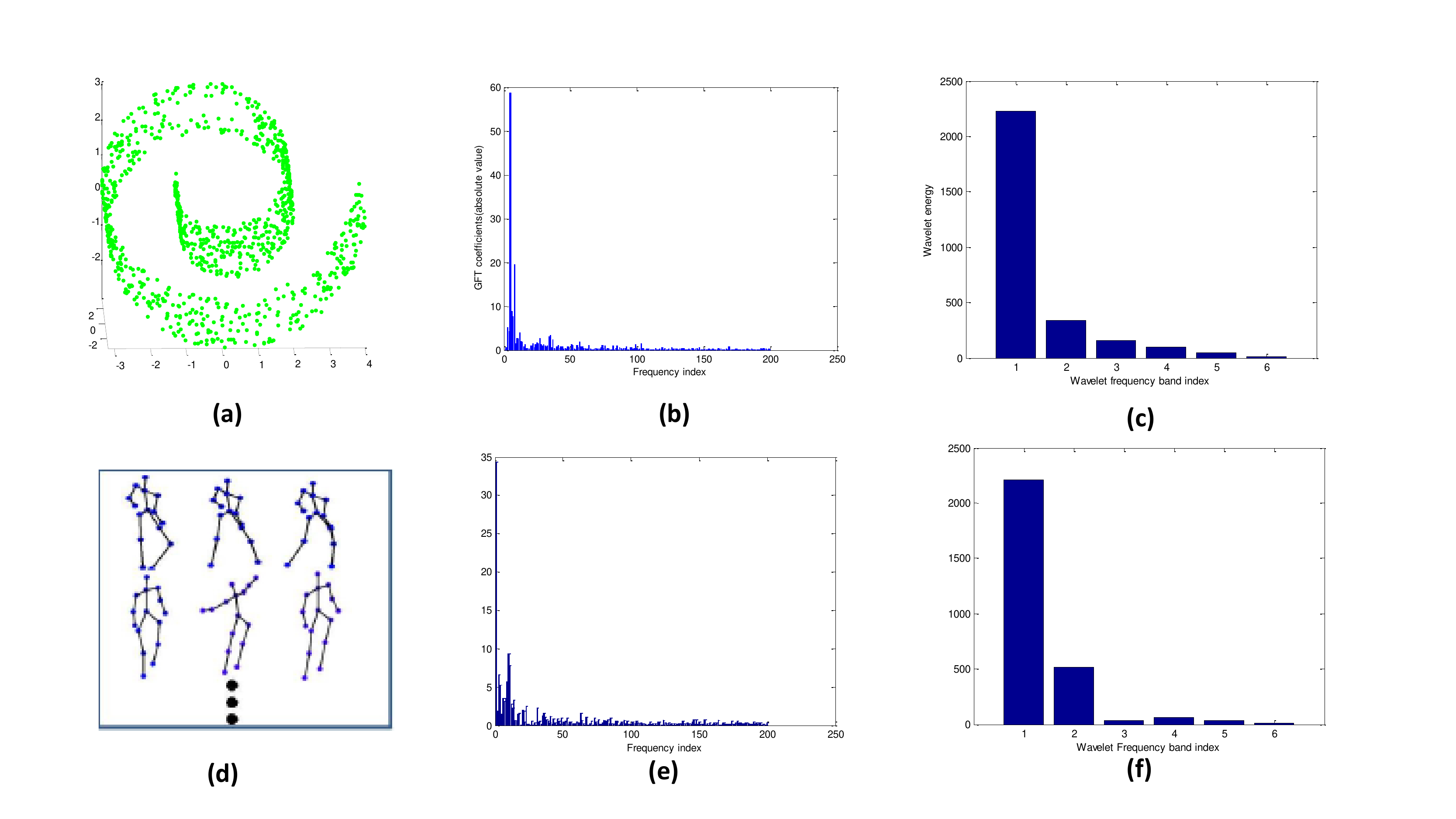}}
  \end{minipage}
\caption{Top raw: Plot of the energy of (a) a noiseless Swiss roll with a hole in (b) the graph Fourier transform (GFT) (b) and in (c) the spectral wavelet domain. Bottom row: (d) Plot of the energy of a noiseless Mocap data (\cite{dianseries}) in the (e) graph Fourier transform (GFT) and in (f) the spectral wavelet domain}
\label{GFT_energy-crop.pdf}
\end{figure}

Our proposed algorithm is motivated by the fact that for smooth manifolds the energy of the manifold coordinate signals is concentrated in the low frequency spectral wavelets, while the noise noise power is spread out in a similar probability distribution across all wavelets bands. 
The former property is illustrated in Figure \ref{GFT_energy-crop.pdf}(b), where it can be seen that most of the energy is concentrated in the GFT coefficients that correspond to the smallest eigenvalues, and similarly (Figure \ref{GFT_energy-crop.pdf}(c)) the energy in each of the wavelet frequency bands for a 6 scale spectral wavelet decomposition can be seen to be concentrated in the low frequency wavelet bands.  

\begin{figure}[htb]
\begin{minipage}[b]{1\linewidth}
\centering
\centerline{\includegraphics[width=16cm]{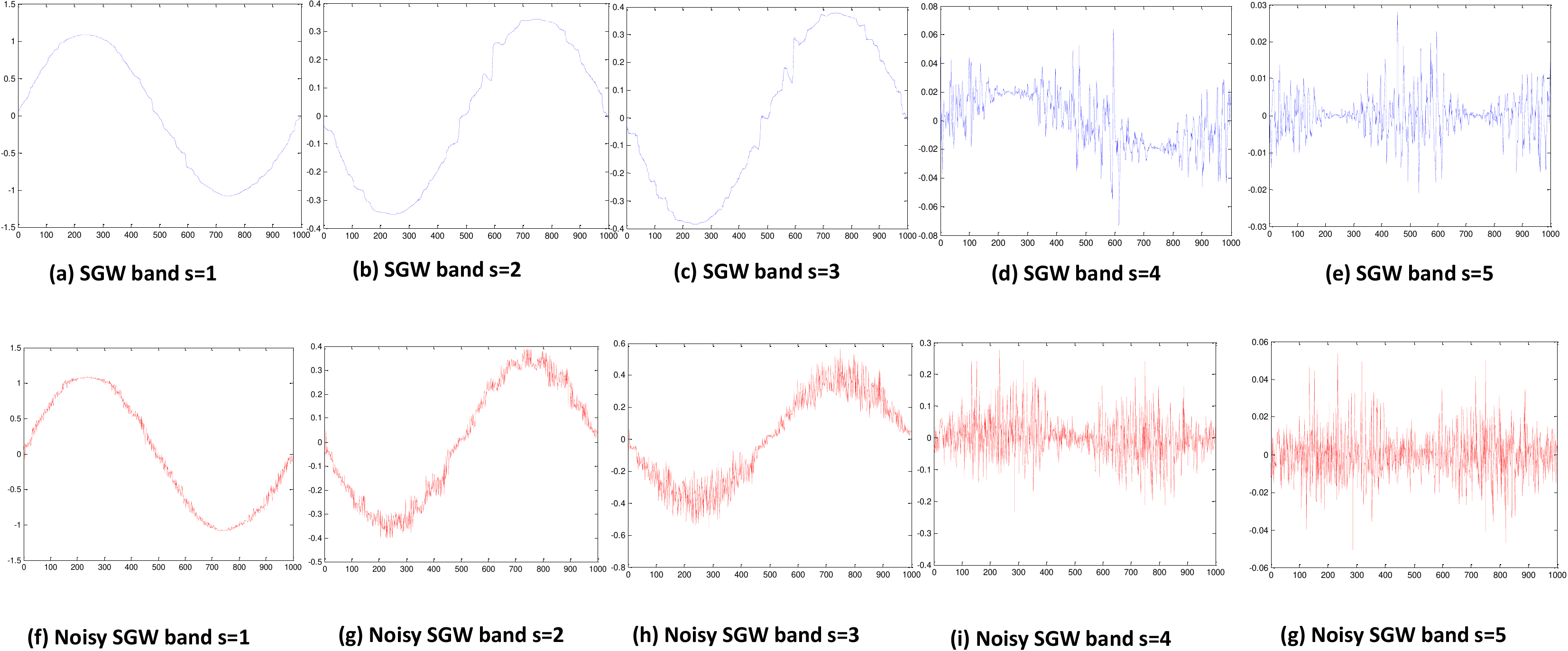}}
  \end{minipage}
\caption{Spectral Graph Wavelets coefficients corresponding to different  frequency bands s of a circle. Figures (a)-(e) plot the SGW bands for an increasing number of frequency band s of a noise-free circle, while Figures  (f)-(g)  plot the SGW bands for a different choice of s for a noisy circle}
\label{SGW_bands}
\end{figure}

It is also important to note the difference between our denoising strategy and shrinkage based methods commonly used in classical wavelet denosing algorithms. In the case of wavelet image denoising, the signals lie on regular grids that are independent of the signal, while in our case, the graph and the noise free signal are closely related by our graph construction. In wavelet denoising for regular signals  we mainly deal with piecewise smooth signals, which lead to a predominantly low frequency signal with localized high frequency coefficients that correspond to discontinuities in the piecewise smooth signal. In contrast, in our graph construction both the domain and the observations depend on the smoothness of the manifold. This has significant implications. For example, if the sampling rate along the manifold varies with the degree of smoothness, we may expect locally smooth behavior of coordinate signals even in areas where the geometry is not as smooth. Thus we do not see SGW domain characteristics similar to what is observed in wavelet domain representation of piecewise smooth regular domain signals (isolated high frequency coefficients). Instead, low frequency spectral wavelet coefficients show locally smooth behavior with respect to the coordinate signals that are spatially connected on the graph, while high frequency spectral coefficients do not appear more predominant than the noise. As an example, Figures \ref{SGW_bands} (a)-(e) show the SGWs in different frequency bands  of a noise free circle. As can be seen, SGWs in the low frequency bands are changing smoothly, while the high frequency bands ($s=4, 5$) are characterized by an oscillatory, non-smooth pattern.  For the noisy SGWs, it can be seen that in the low frequency wavelet bands the power of the true signal is much greater than the noise power, while in the high frequency bands the noise power dominates,  thus making it much harder to separate noise from the signal content (which is different than the case of signals in regular domain). This leads to the approximation of the noise free signal by retaining the low frequency wavelet coefficients and discarding the high frequency wavelet coefficients.

Based on these properties, denoising is performed directly in the spectral graph domain, by retaining all wavelet coefficients that correspond to the low frequency wavelets bands $s \geq s'$,  and discarding all wavelet coefficients in high frequency bands above $s< s'$.

We summarize the proposed denoising algorithm for smooth manifolds in the pseudo code shown in Algorithm \ref{tab:MFD non-iterative}. This approach has several attractive features, in particular, it is  
\begin{description}
 \item (i)  non-iterative, i.e., denoising is performed directly  in the spectral graph wavelet domain in one step. 
 \item (ii) robust against a wide range of $k$ values chosen for nearest neighbor assignment on the graph. 
 \item (iii)  computationally efficient, as the computational complexity is $O(ND)$. 
\end{description}

\begin{algorithm} 
\KwData{The data set $\mathbf{\tilde{x}}= [ \tilde{f_{1}},  \tilde{f_{2}},...,\tilde{f_{D}}]^{t}$, $k$ nearest neighbors on the graph, m - order of Chebyshev polynomial approximation}
Construct an undirected affinity graph $\mathbf{W}$, using Gaussian weights as in (\ref{gaussian_weights}), and construct the Laplacian  $\mathbf{L}$ from $\mathbf{W}$. \;
\For{$r\leftarrow 1$ \KwTo $D$}{
Assign the corresponding coordinate values  $\tilde{f_{r}}$  to its corresponding vertex on the graph. \;
Transform the noisy coordinate signal using SGW defined on $\mathbf{L}$. \;
	Retain all scaling coefficients and all low pass frequency wavelet coefficients which correspond to the largest $s\geq s'$ scales, for which the total accumulated energy is above threshold $E_{thresh}$. Discard all wavelet coefficients above  scale $s< s'$.   \;
       Take the inverse spectral wavelet coefficients of each of then proceed wavelet coefficients.
}
 \KwResult{The reconstructed manifold points $\mathbf{\hat{x}}$.}
 \caption{Manifold Frequency Denoising (MFD) Algorithm}
 \label{tab:MFD non-iterative}
\end{algorithm}

\section{Experimental Results}
\label{sec:results}

\begin{figure*}\begin{minipage}[b]{1\linewidth}
\centerline{\includegraphics[width=12cm]{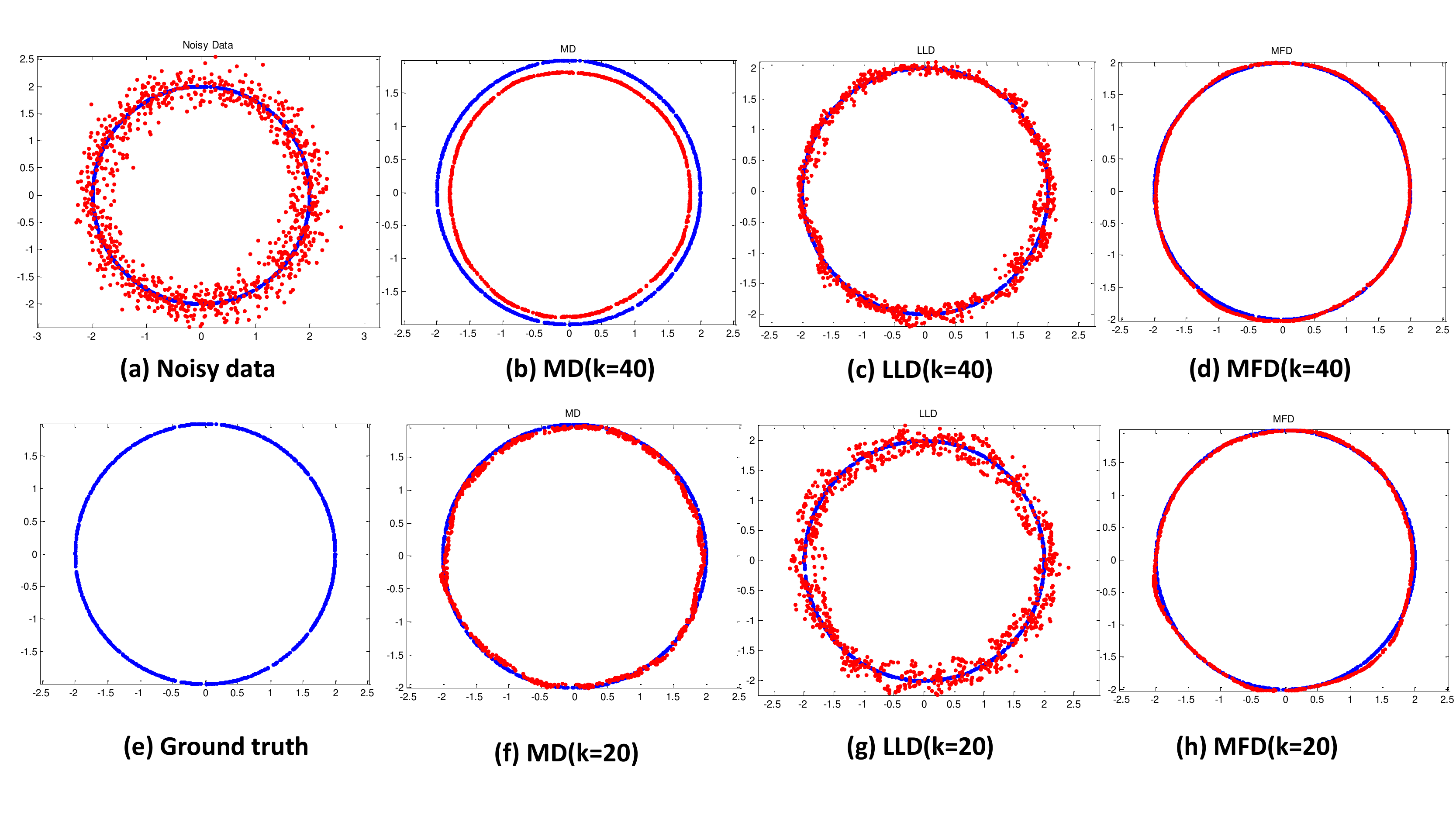}}
\medskip
\end{minipage}
\caption{Experimental results on a circle: (a) Noisy circle (noise shown in red (b) Results with MD  (c) Results with LLD  (d) Results with MFD (e) Ground truth (f) Results with MD (g)  Results with LLD (h) Results with MFD}
\label{Circle_final.pdf}
\end{figure*}

\begin{figure}[htb]
\begin{minipage}[b]{1\linewidth}
\centerline{\includegraphics[width=12cm]{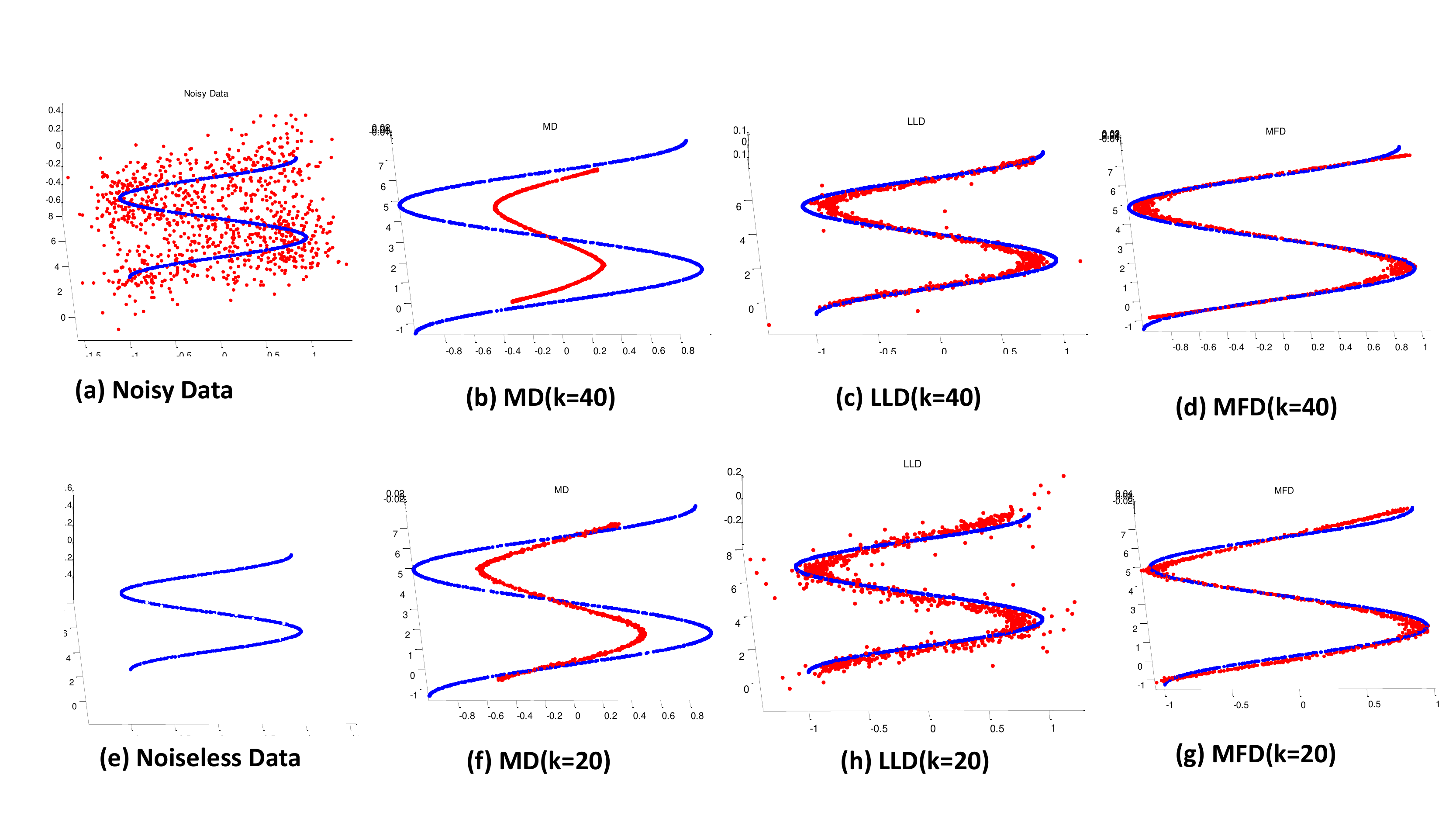}}
\medskip
\end{minipage}
\caption{Experimental results on a sinus function embedded in high dimension:  comparison with different number of $k$ nearest neighbors (a) Results with MD  (b) Results with LLD  (c) Results with MFD  (d) Results with MD (e)  Results with LLD (f) Results with MFD}
\label{Final_sinus_high_dim.pdf}
\end{figure}

\begin{figure*}[t]
\vspace{.10in}
\includegraphics[width=1\linewidth]{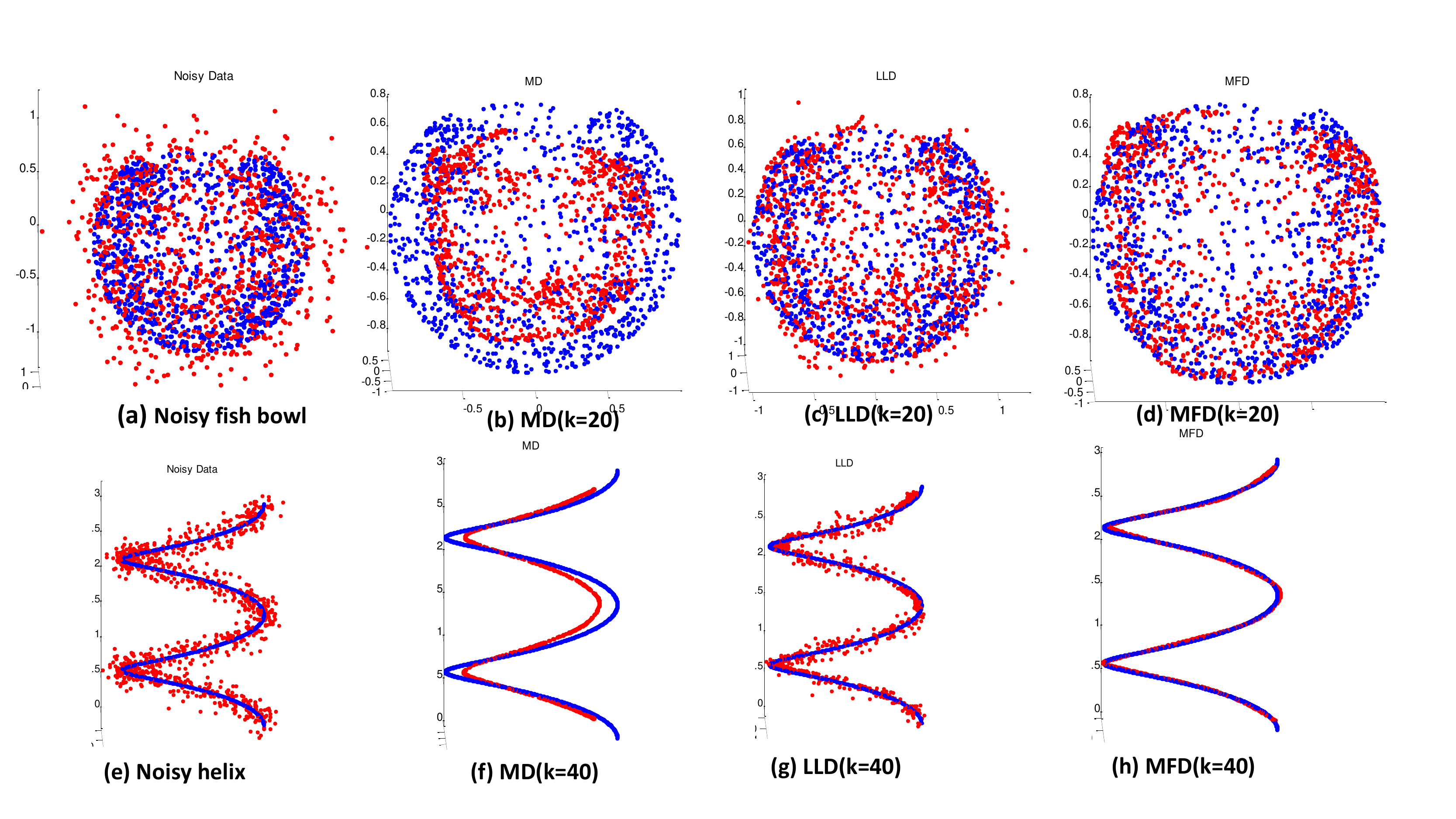}
\caption{Experimental results on a helix and fish bowl manifolds: (a) Noisy fish bowl (noise shown in red color (b) Results with MD  (c) Results with LLD  (d) Results with MFD (e) Noisy helix (noise shown in red color) (f) Results with MD (g)  Results with LLD (h) Results with MFD}
\label{fish_bowl_helix_final.pdf}
\end{figure*}

\begin{figure*}[t]
\vspace{.10in}
\includegraphics[width=1\linewidth]{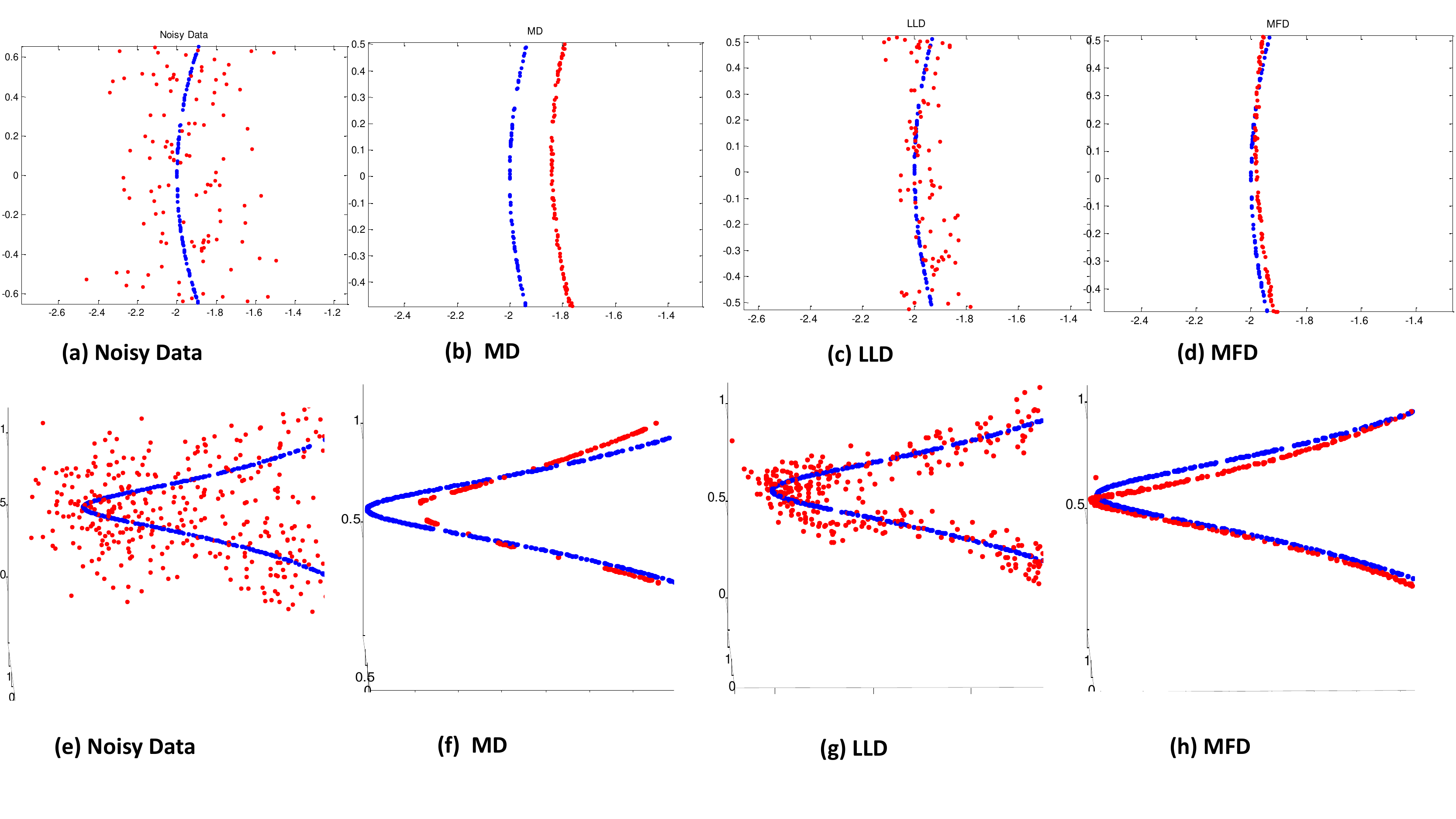}
\caption{Zoom into the the circle and the helix manifolds.  (a) Noisy circle (noise shown in red color, ground truth in blue) (b) Results with MD  (c) Results with LLD  (d) Results with MFD (e) Noisy helix (noise shown in red color, ground truth in blue) (f) Results with MD (g)  Results with LLD (h) Results with MFD}
\label{Zoom_in_circle.pdf}
\end{figure*}

\begin{figure}[htb]
\begin{minipage}[b]{1.2\linewidth}
\centerline{\includegraphics[width=14cm]{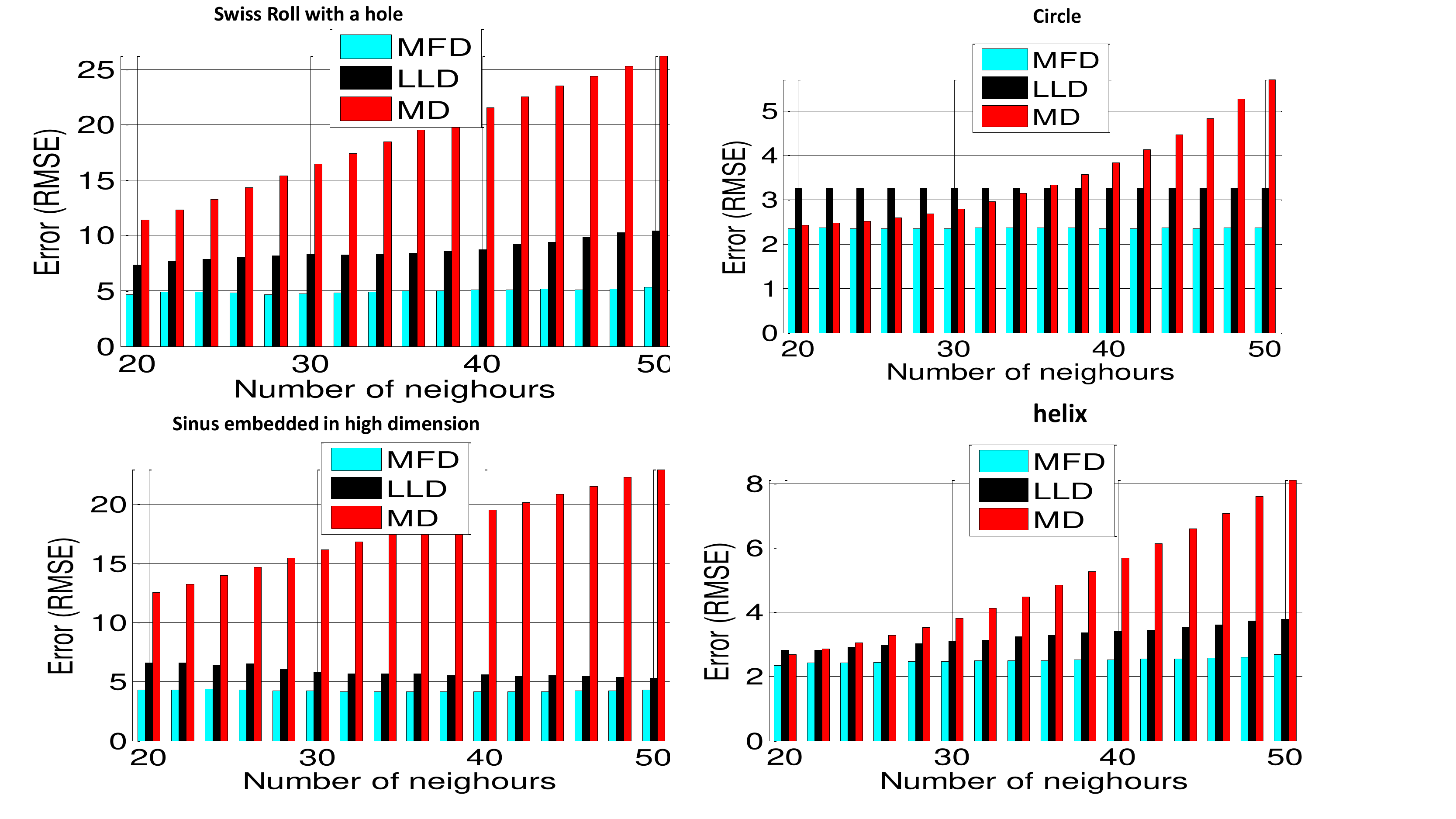}}
\medskip
\end{minipage}
\caption{Experimental results evaluation of the RMSE reconstruction error of the noisy manifolds Swiss roll with a hole, circle, sinus embedded dimension $D=200$, and a helix using different selection of $k$ nearest neighbor}
\label{k_nearest_evaluation}
\end{figure}

We present experimental results with a variety of manifolds, including ones with complex geometric structure such as fish bowl and a Swiss roll with a hole. In addition, we experimented with a sinus function which was embedded  in  high dimensional space of $D=200$. All manifolds were sampled using a uniform distribution with $N=1000$  samples, which were contaminated with isotropic Gaussian noise in all dimensions. In the results shown, the sinus and fish-bowl were contaminated with noise of variance of $0.2$, and the circle, helix and Swiss roll with a hole with variance $0.1$. 
We used $s=5$ wavelet  decomposition levels, and retained all low pass frequency wavelet coefficients which correspond  to the largest $s\geq s'$ scales above total accumulated energy threshold $E_{thresh(var=0.1)}$ for noise variance equal to $0.1$, and the largest scales $s\geq s'$ above total accumulated energy  threshold $E_{thresh(var=0.2)}$ with variance 0.2. The order of the Chebyshev polynomial approximation used was $k/2$ for a $k$ nearest neighbor graph in order to process the manifold locally,  via the approximation of the spectral wavelet coefficients. 

For comparison and evaluation with the state of the art in manifold denoising, we compared our approach to MD (\cite{MDhein} and LLD (\cite{Gongdenosie}). In the case of MD, we found that  a small number of iterations produced better results, while a larger number of iterations produced severe over-smoothing, and thus we fixed the number of iterations to 3 and used the best results.

The experimental results, shown in Figures  \ref{Circle_final.pdf}, \ref{Final_sinus_high_dim.pdf}, and \ref{fish_bowl_helix_final.pdf}, demonstrate that our method significantly outperforms the state of the art, and produces a smooth reconstruction that is faithful to the true topological structure of the manifold. 
In the case of the MD and LLD methods, we observe that for the more complex manifolds or under high noise levels, either the global or local geometric structure was severely distorted. More specifically, MD tends to over-smooth the data (see for example Figures \label{Final_sinus_high_dim}(b) and \label{fish_bowl_helix_final.pdf}(b)), especially for manifolds with complex geometry, while LLD deviates around the mean curvature of the manifold but often fails to produce a smooth reconstruction. This can also be visualized in Figure \ref{Zoom_in_circle.pdf}, where we zoom into the denoising results obtained by MFD and the competing methods. As can be seen, MFD provides a locally smooth reconstruction which also preserves the global  manifold structure, while the competing methods suffer from the limitations previously mentioned.   
We also performed quantitative analysis and compared the reconstruction error in terms of the root mean square error (RMSE) of the denoised manifolds in comparison  to the ground truth. The comparison in Figure \ref{k_nearest_evaluation} shows the performance of MFD, LLD and MD for even numbers of $k$ nearest neighbors graph selection parameter between 20 to 50, for the Swiss role with a hole, circle, sinus embedded in high dimensional space, and a helix. The comparison results in  Figure \ref{k_nearest_evaluation} show that our method is robust against a wide range of $k$ nearest neighbor graph selections, thanks to the multi-scale properties of the Spectral Wavelets. Quantitatively, MFD significantly outperforms LLD and MD for a wide range of $k$ nearest neighbors selection.

 \subsection{Experimental Results on local tangent space estimation} 

Local tangent space estimation is a fundamental step for many machine learning and computer vision applications, which can be severally distorted in the presence of noise. In this section we perform evaluation on local tangent space estimation on noisy manifolds, and show the effect of MFD denoising in terms of the local tangent space estimation error using  popular approaches such as local PCA (\cite{Zhang05}) or Tensor Voting (\cite{Mordohai10}). To test the denoising effect using MFD, we add different amount  of Gaussian noise to a sphere sampled with $N=1000$ points and compare the local tangent space estimation error before and after MFD denoising using Local PCA and Tensor Voting. 
The variance of Gaussian noise was tested in a wide range of parameters shown in Table \ref{tab:local_tangent_estimation}. Local PCA is estimated using $k \in    \left \{ 20,30,40,50,60,70,80 \right \}$  and the scale in Tensor Voting is tested in $ \left \{ 0.1, 0.3, 0.5, 0.7 \right \}$. For each method, we report the best results obtained for the range of parameters  tested. The experimental results of Table \ref{tab:local_tangent_estimation} provides a comparison in terms of the average local tangent space angular error, which is computed using the ground truth normal for all points on the manifold. 
\begin{table} \centering \begin{tabular}{l c | c c c c |c c c c c c c c | } \toprule & \multicolumn{5}{ c |}{Before denoising}  & {  After MFD }\\ \hline
\cmidrule(l){1-10}   $\sigma$ (noise variance)  & 0 & 0.05 & 0.1  & 0.2 & 0.3 &  0.05 & 0.1 & 0.2 & 0.3 \\ \midrule Tensor Voting & \textbf{0.005} & 2.8 & 3.6    & 14.8 & 25.8 & \textbf{1.9} & \textbf{2.1} & 2.9 &\textbf{ 3.4} \\ \midrule
Local PCA        & 1.6    & \textbf{2.5} & \textbf{3.1} & \textbf{6.5} & 10.4 & 2.2 & 2.4 & \textbf{2.7} &  \textbf{3.4}    \\ \midrule \bottomrule \end{tabular}
\caption{Local tangent space estimation average angular error on a sphere using Tensor Voting and local PCA, before and after denoising using MFD method} \label{tab:local_tangent_estimation} \end{table}

The experimental results show that MFD denoising significantly reduces the local tangent space estimation error, especially for mid to high levels of noise (0.1,0.2,0.3). It is also interesting to note that in the noise free case Tensor Voting performs significantly better then local PCA and even converges to zero in the case of manifolds with constant curvature. In the case of mid-to high noise levels, local PCA is less sensitive to noise then Tensor Voting. However both methods show gross errors for higher levels of noise, such that the local tangent space information is severely distorted. Using MFD to denoise the data prior to local tangent space estimation allows to obtain meaningful local structure estimation.

\subsection{Experimental Results with real datasets} 

For experiments with real data-sets, we tested our method on the CMU Motion capture data set, which is a dataset of human motion sequences,  and the Frey faces data-set (\cite{freyref}), a face-expression data-set. The Frey face dataset consists of low resolution faces with dimension D=560,  and in the CMU Motion capture data set we choose 10 mixed sequences from subject 86 where the dimensionality of the data is 62. For the CMU data-set,  in order to perform evaluation in a strictly unsupervised framework, we remove the temporal information from the data. Thus the data provided corresponds to static information which is then is contaminated using Gaussian noise of variance 0.1 in all dimensions. We test our method on the corrupted sequences and compare to the MD and LLD methods. The experimental results in in Table \ref{tab:real_data_mocap} provides an evaluation in terms of the average RMSE error. The  are tabulated in table \ref{tab:real_data_mocap},  where the results shown are the average error obtained in all dimensions and for all sequences. The experimental results obtained using MFD shows significant improvement over LLD and MD.

\begin{table}
\begin{center}
\tabcolsep=0.3cm
\begin{tabular}{|l|c|c|c|}
\hline
\textbf{Data/Method} &  MD  &  LLD & MFD  \\
\hline\hline
CMU MoCap &	11.84	&   7.35 & \textbf{3.42} \\
\hline\hline
Frey face datasets &	109.1	&  62.2 & \textbf{51.2} \\
 & & &  \\
\hline
\end{tabular}
\end{center}
\caption{RMSE average error reconstruction results on motion capture data and Frey face datasets}
\label{tab:real_data_mocap}
\vspace{-5 mm}
\end{table}

\section{Conclusions and Future work}
\label{sec:conclusions}

We have presented a new framework for manifold denoising which simultaneously operates in the vertex and frequency graph domains by using spectral graph  wavelets. The advantage of such an approach is that it allows us to denoise the manifold locally, while taking into account the fine-grain regularity properties of the manifold. Our approach is based on the property that the energy of a smooth manifold concentrates in the low frequency of the graph, while the noise effects all frequency bands in a similar way. 
The suggested MFD framework also possesses additional appealing properties: it is non-iterative and has low computational complexity, as it scales linearly in the number of points for sparse data, thus making it attractive to be used for large scales problems. 

Our current strategy for denoising is based on setting high frequency bands to zero,  which is different from the shrinkage-based methods commonly used in classical wavelet denoising algorithms. While wavelet denoising for regular signals is mostly concerned with piecewise smooth signals that are independent of the regular grid,  in our case, both the domain and the observation depend on the smoothness of the manifold. Moreover, irregular domains can potentially have different characteristics. For example, in some cases locally smooth behavior of the coordinate signals can occur even in areas where the geometry is not as smooth. Our theoretical and experimental study corroborate that for a smooth manifold, the energy  is concentrated in the low frequencies of the graph,  which motivates our strategy to truncate the high frequencies. 

Experimental results on manifolds with complex geometric structure show that our approach significantly outperforms the state of the art, and is robust to a wide range of parameter selection of $k$ nearest neighbors on the graph. 
There are also many other possible future directions for our denoising approach.  

From a practical aspect, MFD provides an effective tool to be used in unsupervised learning applications as a  fast, non-iterative  and efficient denoising method. Moreover, it does not require the knowledge of the intrinsic dimensionality of the manifold, which in many cases is not known in advance or difficult to estimate from the noisy data. It can also be useful in a semi- supervised or supervised problems, in which case the graph is not noisy, thus potentially the denoising process of the coordinates becomes much easier to perform.  
On the other hand, the fact that the behavior of the graph frequencies is primary determined  by the choice of the graph construction leaves many possible theoretical and practical future explorations of the MFD approach by using different graph constructions. For example, a possible limitation of the current suggested denoising algorithm is when the curvature  is changing very rapidly, in which case the initial graph construction may not be sufficiently reliable. This limitation may be solved by using curvature or tangent information, and remains open for future research. Therefore,  a further investigation of how the underlying graph construction affects the spectral transform properties is one of the important future research directions for this framework. Some other promising directions include designing probabilistic modeling in the spectral wavelet domain for smooth manifolds,  addressing the case of non-Gaussian noise, and developing tight bounds for the decay of spectral wavelet at higher frequencies.

\bibliographystyle{plain}
\bibliography{RefJ}

\vskip 0.2in

\end{document}